\documentclass[twocolumn]{svjour3}

\usepackage{framed,multirow}

\usepackage{amssymb}
\usepackage{latexsym}

\usepackage{lineno}

\usepackage{url}
\usepackage{xcolor}
\definecolor{newcolor}{rgb}{.8,.349,.1}

\usepackage{amsmath}
\usepackage{dsfont}
\usepackage{breqn}
\usepackage{multirow}
\usepackage{array}
\usepackage{booktabs}
\usepackage{hhline}
\usepackage{xspace}
\usepackage{arydshln}
\usepackage{hyperref}
\usepackage{graphicx}
\usepackage[numbers]{natbib}

\newcommand{\miniTitle}[1]{\vspace{0.2cm}\noindent\textbf{#1}}
\newcommand{\com}[1] {}

\newcommand{\citenameref}[1]{\citeauthor{#1} \cite{#1}}

\def\makeheadbox{\relax}

\begin{document}

\journalname{Machine Vision and Applications}
 
\title{A generalizable approach for multi-view 3D human pose regression}
\author{Abdolrahim Kadkhodamohammadi \and Nicolas Padoy}

\institute{ICube, University of Strasbourg, CNRS, IHU Strasbourg, France \email{kadkhodamohammad@unistra.fr \and npadoy@unistra.fr}}

\maketitle

\begin{abstract}
Despite the significant improvement in the performance of monocular pose estimation approaches and their ability to generalize to unseen environments, multi-view approaches are often lagging behind in terms of accuracy and are specific to certain datasets. This is mainly due to the fact that (1) contrary to real world single-view datasets, multi-view datasets are often captured in controlled environments to collect precise 3D annotations, which do not cover all real world challenges, and (2) the model parameters are learned for specific camera setups. To alleviate these problems, we propose a two-stage approach to detect and estimate 3D human poses, which separates single-view pose detection from multi-view 3D pose estimation. This separation enables us to utilize each dataset for the right task, i.e. single-view datasets for constructing robust pose detection models and multi-view datasets for constructing precise multi-view 3D regression models. In addition, our 3D regression approach only requires 3D pose data and its projections to the views for building the model, hence removing the need for collecting annotated data from the test setup. Our approach can therefore be easily generalized to a new environment by simply projecting 3D poses into 2D during training according to the camera setup used at test time. As 2D poses are collected at test time using a single-view pose detector, which might generate inaccurate detections, we model its characteristics and incorporate this information during training. We demonstrate that incorporating the detector's characteristics is important to build a robust 3D regression model and that the resulting regression model generalizes well to new multi-view environments. Our evaluation results show that our approach achieves competitive results on the Human3.6M dataset and significantly improves results on a  multi-view clinical dataset that is the first multi-view dataset generated from live surgery recordings.

\end{abstract}


\section{Introduction}
\label{intro}

\begin{figure*}[tb]
\centering
\setlength{\tabcolsep}{0.5pt}
\renewcommand{\arraystretch}{1}
\begin{tabular}{c c c}
\includegraphics[width=0.33\textwidth]{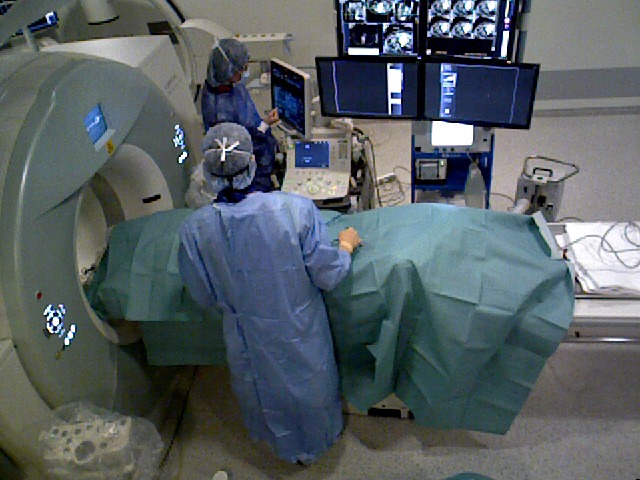} &
\includegraphics[width=0.33\textwidth]{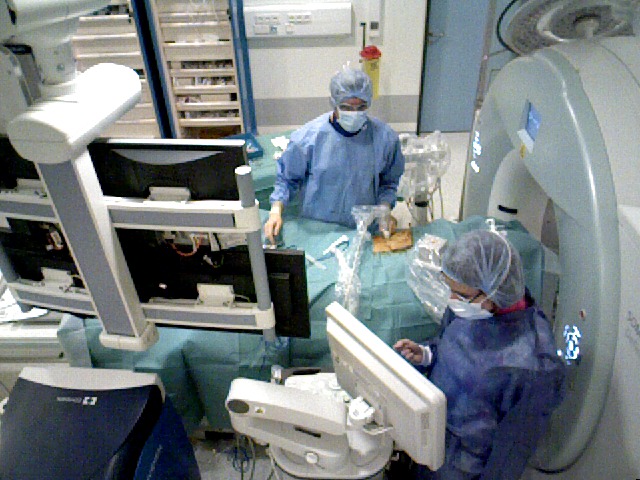} &
\includegraphics[width=0.33\textwidth]{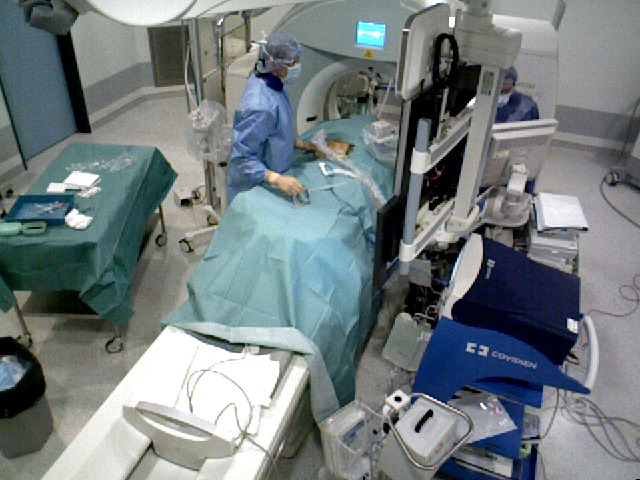} 
\end{tabular}
\caption{A set of images captured by a multi-view camera system at the same time step. Even though some body parts are occluded in one view due to self- or object-occlusion, they might appear in other complementary views.}
\label{fig:mvFrame}
\end{figure*}

Single-view human detection and body pose estimation have enjoyed a great deal of attention over the last decades in the field of computer vision because of their importance for various applications, ranging from activity recognition to human computer interaction. More recently, the emergence of deep learning has pushed the boundaries in many fields, including computer vision. The combination of deep learning with the availability of large datasets, such as {\it MPII Pose}~\cite{andriluka_cvpr2014} and {\it MS COCO}~\cite{MSCOCO}, has spawned many promising approaches for single-view human detection and pose estimation \cite{wei_cvpr2016,newell_nips2017,cao_cvpr2017}. But the presence of clutter and occlusions degrades their performance. Capturing an environment from complementary views permits to reduce the risk of occlusions, especially in busy environments, as shown in Figure~\ref{fig:mvFrame}. In addition, the availability of calibrated multi-view data greatly facilitates the process of lifting 2D scenes into 3D, which is important for many applications such as augmented reality.

Despite the inherent benefits of capturing an environment from multiple views, multi-view approaches have not achieved the same level of maturity as compared to single-view approaches, mostly due to two reasons: firstly, multi-view datasets are generally recorded in controlled environments in order to use motion capture systems to acquire precise 3D ground truth location data. This removes the need for the tedious and error-prone manual annotation of the abundant number of frames coming from all views for generating ground truth 3D poses. Even though there are large multi-view datasets such as {\it Human3.6.M} \cite{ionescu_pami2014} and {\it HumanEva} \cite{sigal_ijcv2009}, the simple backgrounds and tight clothes required by motion capture systems make these datasets trivial for 2D pose estimation methods. Monocular pose estimation approaches report low 2D body part localization errors even without finetuning \cite{chen_cvpr2017,martinez_iccv2017}. For these reasons, single- and multi-view pose estimation models trained on datasets captured in such controlled laboratory environments do not generalize well to real world data, which is often visually much more complex due to occlusions, clutter and the presence of multiple persons in the scene. 
Secondly, current multi-view approaches \cite{sigal_ijcv2012,dogan_iet2017,pavlakos_cvpr2017_2} learn model parameters that are specific to each multi-view camera setup. In other words, to apply these approaches on a {\it new} multi-view scenario, it is required to collect new annotated data that includes both multi-view images and their corresponding 3D ground truth poses for the same camera setup. On the one hand, generating synthetic datasets for these approaches would require not only the generation of 3D body poses, but also of photo-realistic rendering of humans with different shapes, textures and backgrounds to allow generalization to the real world, which is not a trivial task. On the other hand, generating such training data using either motion capture systems or manual annotations, especially in the case of data-hungry deep learning methods, is not always feasible in uncontrolled environments and very tedious. We therefore propose an approach that benefits from existing multi-view datasets to perform multi-view 3D pose estimation in new multi-view setups. 


Our approach formulates the problem of multi-view 3D pose estimation in a two-step framework: (1) single-view pose detections and (2) multi-view 3D pose regression. We separate these two steps for two reasons. First, we can better exploit available single-view and multi-view datasets for the right task. Single-view datasets, such as  {\it MPII Pose}~\cite{andriluka_cvpr2014} and {\it MS COCO}~\cite{MSCOCO}, include diverse and challenging frames from everyday activities or movies originating from amateur to professional recordings. Therefore, models trained on these datasets can better cope with real world challenges and generalize to new environments.
But, these single-view datasets are lacking 3D annotations, contrary to multi-view datasets, which often come with accurate 3D body poses. As these are however much simpler for the task of 2D pose estimation \cite{chen_cvpr2017,martinez_iccv2017}, researchers have proposed methods to jointly use both single- and multi-view datasets in order to construct more robust 3D pose estimation models from multiple views \cite{amin_pr2014,belagiannis_mva2016}.
Changes in camera setups however require the retraining of the model on training data from the {\it same} camera setup. This strictly limits the deployment of the models to environments where such training data exists. 
The second reason for our two steps approach is that we can better generalize to new multi-view environments by assuming that lifting 2D body poses into 3D is independent of the images given the 2D pose detections. This assumption implies that we do not need to collect 2D image data for training the 3D regression function and that any set of plausible 3D body poses can be used instead by computing body pose projections into 2D.

To learn a multi-view 3D regression function, we propose a method that relies on a multi-stage neural network. The input of this network is a set of corresponding multi-view 2D detections for each individual person. At test time, they are collected using a state-of-the-art single-view detector. 
We assume that the camera system is fully calibrated and can therefore use epipolar geometry to establish the multi-view correspondences per person. 
This process also allows us to detect the number of persons per multi-view frame\footnote{We define a multi-view frame as the set of all images captured from all views at the same time step.}. This is in contrast to current multi-view RGB approaches, which tackle either single-person scenarios \cite{gall_ijcv2010,hofmann_ijcv2011} or multi-person scenarios where the number of persons is known {\it a priori} \cite{luo_icpr2010,belagiannis_eccv2014}.

The proposed network consists of a series of blocks of fully-connected layers with intermediate supervision at each block. The input to each block is the raw network input, i.e. the concatenated 2D poses, and the output from previous block if it exists. The network can therefore build a high dimensional function and refine the output of the previous block to achieve a more reliable regression function. In order to generalize to new multi-view setups, we do not use images during training but construct training data solely by projecting Human3.6M's 3D poses. We use Human3.6M because it is the largest publicly available multi-view dataset and it includes men and women of different sizes.   
The projected 2D poses are generated according to the camera parameters used at test time. In practice, 2D poses are detected at test time using a 2D pose detector that may be noisy and inaccurate. In order to cope with these inaccuracies, we propose to perturb during training the 2D locations of the body joints by random noise that is generated based on the characteristics of the 2D detector. We also propose to incorporate a detection confidence for each body joint, computed based on the amount of noise added during training. This provides a representation for the detection confidence generated by the detector at test time. Therefore, the approach can take into account not only joint locations, but also detection precision to build a robust regression function.  

We use two datasets to perform quantitative and qualitative evaluations and compare with state-of-the-art results on these datasets. 
We first report results on the Human3.6M dataset \cite{ionescu_pami2014} to characterize the properties and the performance of our approach. 
This dataset includes recordings of several actions performed by professional actors of different genders. This dataset has been recorded by a fully calibrated four-view camera system and a motion capture system to collect ground truth 3D positions of the body joints. We also evaluate our approach on a challenging multi-view dataset \cite{kadkhoda_wacv2017} to show the generalization ability of our approach. This dataset is generated from real surgery recordings obtained in an operating room (OR) using a three-view camera system and hence is called {\it Multi-view OR} (MVOR) in the following.
Our approach improves 3D body part localization on Human3.6M and significantly reduces the localization error on the multi-view OR dataset without using any training data from this dataset.
 
The main contributions of the paper are twofold. First, we present a simple and yet accurate multi-view 3D pose estimation approach that can generalize well to new multi-view environments. In contrast to current state-of-the-art methods, the approach exploits an existing multi-view dataset to build models for new multi-view environments without any need for new annotation. Second, this is the first multi-view RGB approach that has been quantitatively evaluated on data captured in an unconstrained environment.

\section{Related Work}

\miniTitle{Multi-view segmentation-based 3D pose estimation.}
\citenameref{hofmann_ijcv2011} use foreground  segmentation to estimate body silhouettes per view. Then, 3D pose candidates are obtained by matching a library of exemplars. Texture information and shape similarity across all views combined with temporal information are used to compute the final 3D poses. Similarly, \citenameref{gall_ijcv2010} propose a two-layer framework that iteratively improves foreground segmentation and retrieved body poses by incorporating both multi-view and temporal information. Other approaches have deployed optical flow estimation \cite{chen_spcs2008}, 2D as well as 3D motion cues \cite{Sundaresan_tip2009} and low-rank multi-view feature fusion combined with sparse spectral embedding~\cite{yu_nc2017} to estimate 3D poses. In contrast to our work, these approaches are only evaluated on single-person datasets. More importantly, it is not always possible to compute foreground in cluttered environments, such as in operating rooms. Therefore, these approaches can only be evaluated on data recorded in environments with simple backgrounds. 

\miniTitle{Multi-view part-based 3D pose estimation.}
Several multi-view 3D pose estimation approaches \cite{burenius_CVPR2013,amin_bmvc2013,amin_pr2014,belagiannis_CVPR2014,belagiannis_mva2016,kadkhoda_wacv2017} have been proposed that rely on a part-based framework \cite{felzenszwalb_pami2010}. This part-based framework provides an elegant formalism to optimize over different potential functions for incorporating image features, multi-view cues, temporal information and body physical constraints.
\citenameref{burenius_CVPR2013} propose an approach that extends pictorial structures \cite{fischler_TC1973,felzenszwalb_ijcv2005} to multi-view and to perform exact 3D inference by using simple binary pairwise potential functions. Instead,  Amin et al. \cite{amin_bmvc2013,amin_pr2014} use 2D inference with more complex pairwise potentials, multi-view cues and triangulation to estimate 3D poses. \citenameref{belagiannis_CVPR2014} have also deployed different pairwise potentials for incorporating both body physical constraints and multi-view features. This approach allows to perform approximate 3D inference by selecting a limited number of hypotheses per individual. This approach has further been extended to incorporate temporal information \cite{belagiannis_eccv2014} and to use a deep neural network based body part detector~\cite{belagiannis_mva2016}. Recently, \citenameref{pavlakos_cvpr2017_2} has used deep neural network to predict body part score maps across all views and then estimated body poses by using a 3D pictorial structures approach.  

In contrast to our work, all these approaches have only been evaluated on datasets recorded in constrained laboratory environments and also require the number of person to be known {\it a priori}. MVDeep3DPS presented in \cite{kadkhoda_wacv2017} is an exception, but this approach relies on multi-view RGB-D input to estimate 3D body poses. Additionally, all these approaches need in general to learn model parameters on data from the same camera setup. Moreover, optimizing these energy functions is demanding, especially in 3D, which makes these approaches not suitable for real-time applications. In our work, we do not require images with pose annotations from the camera setup used at test time and learn model parameters by using existing datasets. Furthermore, our approach performs both human detection and pose estimation. As our regression function uses a multi-layer neural network, it runs in super real-time on a single consumer GPU card.    


\miniTitle{Single-view 3D pose estimation.}
Recently, many deep learning based approaches have been proposed to directly regress body poses in 3D from a monocular image or an image sequence. 
\citenameref{pavlakos_cvpr2017} use a stack of a fully convolutional network \cite{newell_eccv2016} to iteratively compute 3D heatmaps per body parts. \citenameref{tekin_bmvc2016} propose to learn an auto-encoder that maps 3D body joints into a high-dimension latent space for discovering joint dependencies and then to learn a convolutinal network that maps an image into this high-dimensional pose space. In \cite{tekin_cvpr2016}, motion compensation is used to align several consecutive frames and construct a rectified spatiotemporal volume that is then fed into a 3D regression function. Other approaches have built deep pose grammar representations \cite{fang_arXiv2017}, skeleton map \cite{wan_arXiv2017} and multitask objectives \cite{rogez_cvpr2017,luvizon_cvpr2018} to enforce more constraints and obtain a more accurate 3D regression function. These approaches are trained on images with accurate 3D ground truth poses. The main issue is that to generate such accurate 3D annotations, motion capture systems are used in controlled laboratory environments with simple backgrounds. Models trained on such image data do not generalize well to real world scenes.

Another line of work relies on two-stage methods, where 2D body parts are first predicted using 2D pose detectors \cite{wei_cvpr2016,newell_eccv2016,cao_cvpr2017} and then 3D body part locations are computed by relying on these predictions \cite{moreno_cvpr2017,chen_cvpr2017,martinez_iccv2017}. In comparison with direct 3D regression approaches, these approaches benefit from the diverse, challenging and real world datasets, e.g. MS COCO and MPII Pose, to train reliable 2D pose detector models that generalize well. To compute 3D body locations, exemplar-based approaches are used by matching lower and upper body parts separately \cite{jiang_icpr2010} and by matching the whole skeleton \cite{chen_cvpr2017}. More recently, \cite{moreno_cvpr2017} proposed to regress from 2D Euclidean distance matrices (EDM) to 3D EDM instead of using traditional 2D-to-3D regression in the Cartesian coordinate system \cite{radwan_iccv2013,ionescu_pami2014}. The regression is performed using a fully convolutional network and 3D poses are recovered via a multidimensional scaling algorithm \cite{biswas_tase2006}. \citenameref{martinez_iccv2017} showed that a simple fully connected network to regress from 2D to 3D outperforms \cite{moreno_cvpr2017} and achieves state-of-the-art results on Human3.6M. We also adopt a two-stage framework in our multi-view approach and use a fully connected network as a 2D-to-3D regression function. The single-view model in \cite{martinez_iccv2017} was however trained on the output of the 2D detector used during test time. In contrast, our approach relies solely on ground truth during training and instead generates training samples that comply with the behavior of the 2D detector used at test time. This is an interesting property of our approach, which enables us to train our network on Human3.6M and test on a completely different multi-view dataset.     

\section{Methodology}

In this section, we present our proposed approach for multi-view 3D pose estimation. We assume that we have a calibrated multi-view system recording an environment from a set of complementary views. Our objective is to detect and predict human body poses in 3D given images captured from all views. In a probabilistic formulation, we want to compute $p(Y,\mathds{X,I})$, the joint distribution over the following three random variables: (1) the 3D body poses
 $Y=(y_1,y_2 ...,y_P)$, where $P$ is the number of body joints and $ \; y_i \in \mathds{R}^3$ is a body joint location in 3D; (2) the 2D body poses $\mathds{X}=(X_1,X_2,...,X_V)$, where $V$ is the number of viewpoints and $X_j$ is the tuple of pixel coordinates indicating the body joints of a 2D pose in view $j$; and (3) all 2D images $\mathds{I}=(I_1,I_2,...,I_V)$, where    $I_j$ is the image taken from the $j^{th}$ viewpoint. Such a formulation makes no limiting assumption and indicates that a 3D body pose is jointly dependent on its appearance in all individual views. However, learning such a model requires collecting training data from the same multi-view setup that we want to apply the model to.

Without loss of generality, we can rewrite the joint probability distribution as: 
\begin{equation}
\centering
p(Y, \mathds{X,I}) = p(Y|\mathds{X,I}) . p(\mathds{X}|\mathds{I}) . p(\mathds{I}).
\end{equation}
To build a multi-view pose estimation approach that can generalize to new environments, we make two conditionally independence assumptions. Firstly, the 3D pose $Y$ is assumed conditionally independent of images $\mathds{I}$ given 2D poses $\mathds{X}$. Obviously, this is not always correct, as one can find different 3D skeletons that have similar 2D projections due to the 3D-2D perspective effect. The likelihood of such cases however degrades dramatically in a multi-view setup, where a working volume has been captured from complementary views. \\
Secondly, we assume that given an image observation for a view $j$, 2D poses in this view are conditionally independent of detections in the other views and other image observations. One can see that this assumption does not hold in case of occlusions. But, we believe that this assumption is reasonable for these three reasons: (1) there exist challenging single-view datasets, e.g. MS COCO and MPII Pose, which can be used to train robust single-view pose detection models; (2) recent deep neural network based approaches have achieved very promising results on unseen data and reliably discriminate occluded joints from visible ones \cite{cao_cvpr2017,newell_eccv2016,newell_nips2017}; and (3) it yields an interesting modeling that allows us to train a 2D pose detector independently. Considering these two assumptions, we can rewrite the joint probability as: 
\begin{equation}
\centering
p(Y, \mathds{X,I}) = p(Y|\mathds{X}) . \prod_{j=1}^V \big(p(X_j|I_j) . p(I_j)\big).
\end{equation}
This equation indicates that a 2D pose detector is applied in each view independently and that the 3D pose regression function is solely dependent upon 2D pose detections.
We model the first term using a multi-view 3D regression function, described in Section~\ref{sec:3dRegFunc}. The input for this function is provided by concatenating 2D detections for each individual person across all views, which is presented in Section~\ref{sec:catDet}. The second term is the single-view pose detector explained next. 



\subsection{Single-view 2D Pose Detector}
\label{sec:svPoseDet}
The relaxation assumption mentioned above allows us to use arbitrary complex models to detect and localize 2D body poses given single-view images. We therefore use the deep convolutional network of \cite{cao_cvpr2017} as single-view pose detector. This approach is currently the state-of-the-art approach for multi-person 2D pose estimation. In addition to its reliable multi-person pose estimation performance, the approach runs in nearly real-time. Given an image, the model generates a set of 2D poses, where each body pose is specified by a collection of 18 body parts. For each body part, the model provides its pixel coordinate and a detection confidence. The confidence values are in range $[0,1]$, where zero indicates undetected body parts.


\subsection{Concatenating Detections Across all Views}
\label{sec:catDet}

\begin{figure}[tb]
\centering
\setlength{\tabcolsep}{1pt}
\renewcommand{\arraystretch}{1}
\begin{tabular}{cc}
\includegraphics[width=0.495\columnwidth]{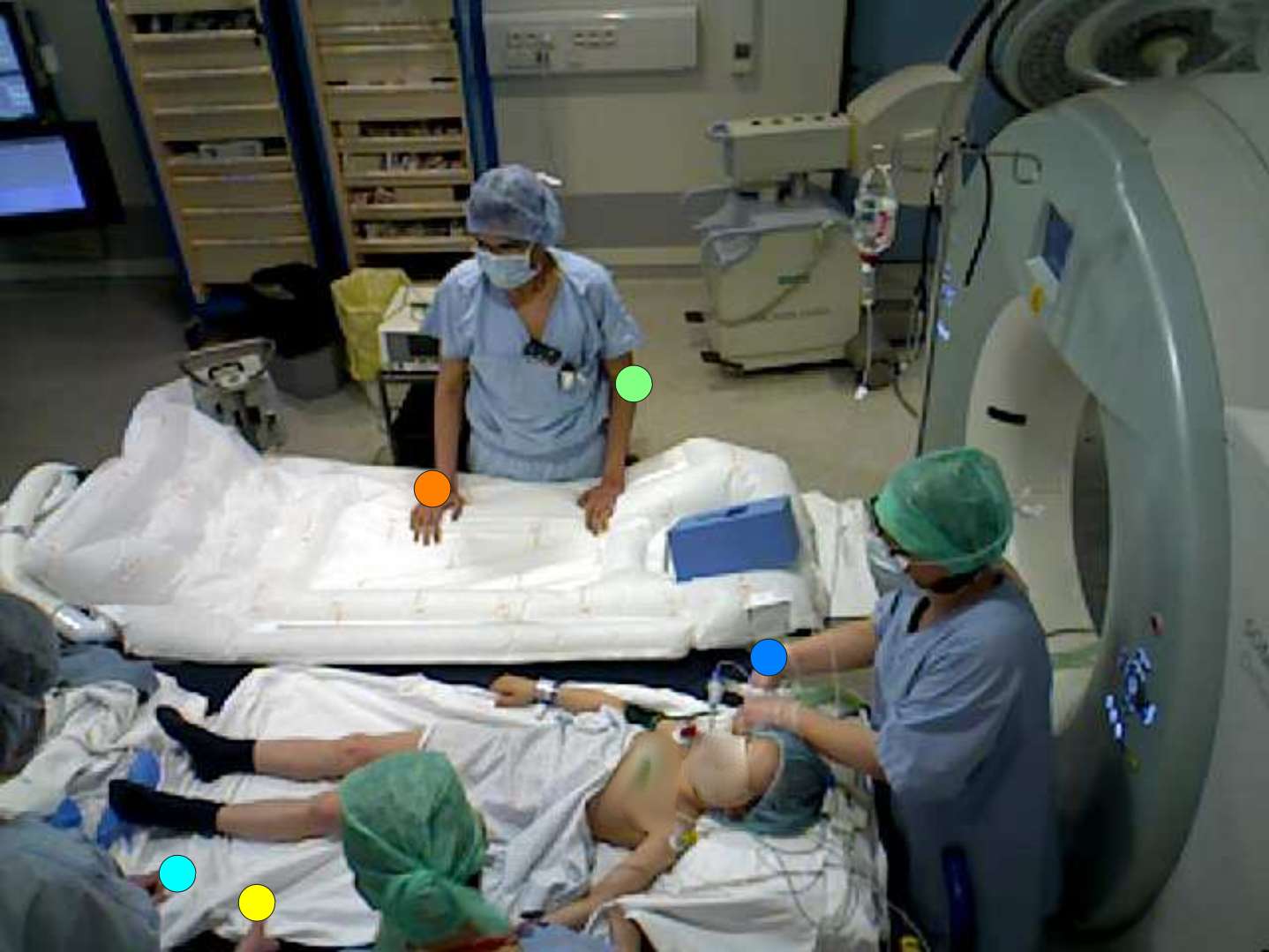} & 
\includegraphics[width=0.495\columnwidth]{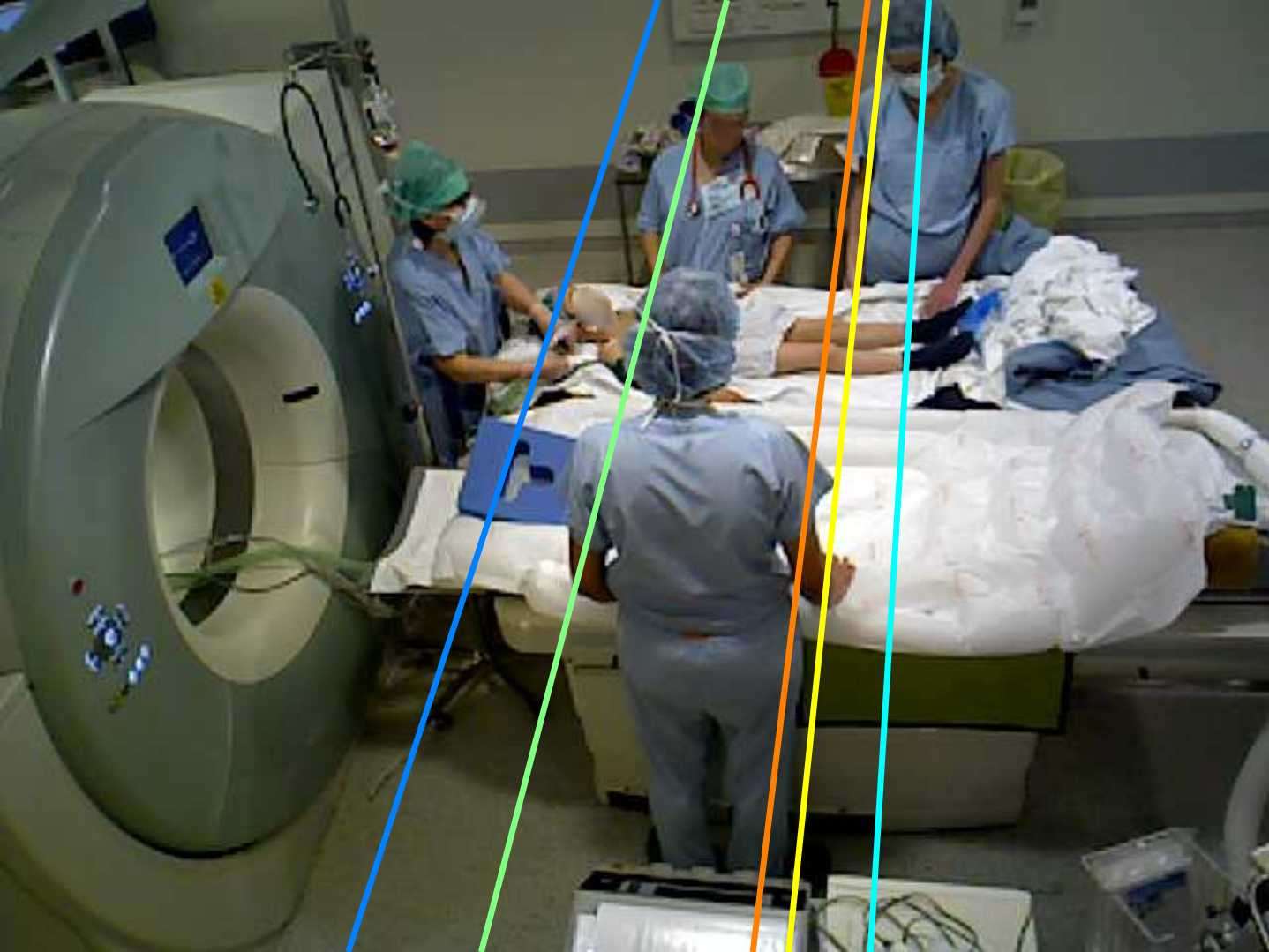}
\end{tabular}
\caption{Person matching using epipolar geometry. A set of points and their corresponding epipolar lines are shown for a pair of images captured from two different viewpoints at the same time step. (Best seen in color)}
\label{fig:rectify}
\end{figure}

Given the detected poses per view, we need to find correspondences across the views. As we assume that the camera system is fully calibrated (i.e. both camera intrinsic and extrinsic parameters are available), we use epipolar geometry to find correspondences \cite{Hartley00}. Let us assume that for each pair of cameras $(C,C^')$ the camera parameters are given with respect to the first one:
\begin{equation}
\centering
C = K[I|\mathbf{0}] \;\; \mathtt{ and }\;\; C^'=K^'[R|\mathbf{t}],
\end{equation}
where $K$ and $K^'$ are camera intrinsic parameters and $[A|\mathbf{b}]$ indicates extrinsic parameters.
 We can compute the fundamental matrix $F$ by:
\begin{equation}
\centering
F = K^{'-T} R K^T [KR^T\mathbf{t}]_{\times},
\end{equation}
where $[\mathbf{b}]_{\times}$ is the skew matrix operator.
The fundamental matrix encapsulates all cameras parameters and allows us to compute the corresponding epipolar line for a point in the other view, as illustrated in Figure~\ref{fig:rectify}.  

Here, we use the fundamental matrix to compute average distances between detected skeletons for all pairs of views. This distance is computed for each possible pair of detections from two distinct views as the average distance between a subset of body joints detected in both skeletons. We collect 2D skeletons for each person across two views by computing the average distances between detected skeletons in one view and the corresponding epipolar lines of  skeletons from the other view and by then finding disjoint pairs of skeletons with the lowest average distance. We exclude pairs for which the average distance is bigger than 20 pixels.
We then use the matched skeletons to establish multi-view correspondences per individual person. One should note that despite the availability of the correspondences, we cannot use triangulation because inaccurate detections lead to high error in 3D and, more importantly, joints might be detected in less than two views, especially in cluttered environments. 
We therefore use a regression function to compute the 3D positions of the body joints.   

To prepare the input for the regression function, we concatenate skeletons across all views. If a person is not detected in a view, we fill the corresponding entry with zeros. Each body part is represented by three channels: two channels indicating pixel location and the third channel indicating the detection confidence. 

\subsection{Training Data Generation}
\label{sec:trainDataGen}

As mentioned in the introduction, we generate training samples by projecting 3D skeletons into 2D. The model can therefore be trained on data generated from existing datasets or any set of valid 3D poses. The projected 2D skeletons are computed based on the camera setup used at test time. Since the single-view 2D pose detector used at test time can provide noisy detections, the model needs to be trained on similar noisy detection data to be able to generalize. We therefore evaluate our 2D pose detector on the Human3.6M dataset, which contains both images and ground truth 2D poses, to characterize its performance. We use these evaluation results to design a normally distributed noise model for each body joint. This noise is used to perturb training data.
We then compute the confidence for the joint as:
\begin{equation}
\label{eq:noiseConf}
\centering
conf = \max(1 - \frac{w}{\lambda . \sigma}, 0),
\end{equation}
where $w$ is the amount of additive noise, which is sampled from a normal distribution with zero mean and standard deviation $\sigma$, and $\lambda$ is a coefficient. We use this coefficient to set the confidence of a joint to zero, i.e. undetected, based on the relative amount of added noise with respect to the standard deviation. 
We use the evaluation results of \cite{cao_cvpr2017} on Human3.6M, presented in Section\ref{sec:2dDectRes}, to set these parameters. 
As shown by the experiments, perturbing trained data and incorporating the confidence value are important for the method to generalize well to unseen data. 


\begin{figure}[tb]
\centering
\setlength{\tabcolsep}{1pt}
\renewcommand{\arraystretch}{1}
\includegraphics[width=0.98\columnwidth]{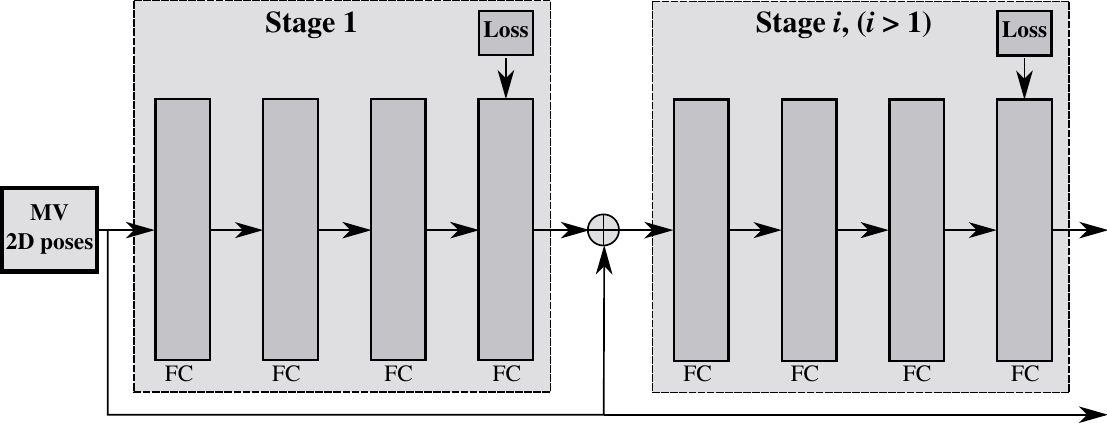} 
\caption{Regression network architecture. The network consists of several stages. Each stage includes four fully connected (FC) layers and intermediate supervision is provided by computing an $L2$ loss at the last FC layer in each stage. The network takes as input a vector of 2D poses concatenated across all views for each individual person, as presented in Section \ref{sec:catDet}.}
\label{fig:networkArch}
\end{figure}

\subsection{Multi-view 3D Regression Function}
\label{sec:3dRegFunc}

As mentioned earlier, the regression function relies solely on the detections provided by the single-view 2D pose detector. In contrast to \cite{tekin_bmvc2016,fang_arXiv2017,luvizon_cvpr2018}, we do not need to model a complex function to directly map image pixel intensities into body part locations in 3D. Similar to \cite{martinez_iccv2017}, we model the 3D regression function using a simple multi-stage multilayer neural network. 

The illustration of the network architecture is shown in Figure~\ref{fig:networkArch}. The network consists of several stages, where each stage is made of four fully connected (FC) layers. The first stage takes the multi-view 2D detections as input, described in Section \ref{sec:catDet}.
Every stage in this network is trained to regress for the desired output. This provides intermediate supervision at each stage and automatically alleviates the problem of vanishing gradient that happens when there are many intermediate layers between the network input and output layers~\cite{cao_cvpr2017}. We can therefore build deep neural networks by stacking several stages. The stage-wise supervision is provided by computing the $L2$ loss between the output of the last layer in each stage and the desired output ($y^*$):
\begin{equation}
\centering
\mathcal{L}_s = \frac{1}{N}\sum_{n=1}^{N} ||y_n^s-y_n^*||_2^2, 
\end{equation}
where $\mathcal{L}_s$ is the average loss computed over all $N$ training samples used in this iteration and $y_n^s$ is the output of the last layer at stage $s$ for sample $n$. The network is optimized by computing the overall network loss as a sum of the losses from all $S$ stages that is defined as:
\begin{equation}
\centering
\mathcal{L} = \sum_{s=1}^S\mathcal{L}_s.
\end{equation}

Since we need to retrain the model for new multi-view setups, we use batch normalization in order to reduce sensitivity to network initialization and learning rate \cite{ioffe_icml2015}. We have also used dropout to avoid overfitting~\cite{srivastava_jmlr2014} and rectified linear units to achieve non-linearity~\cite{nair_icml2010}.

\section{Experiments}
\label{sec:exp}

In this section, we present the evaluation on two multi-view datasets and compare with state-of-the-art results.

\subsection{Implementation Details}
We implement our approach using TensorFlow \cite{tensorflow}. In each stage of the network, the size of the first and last layers are set based on the input and output dimensions and the size of the intermediate layers are set to 1024.  
Our network is trained using the Adam optimizer. We set the starting learning rate to 0.001 and use exponential decay. The batch size is set to 512 and we train our network for 200 epochs. We observe that the performance of the network reaches a plateau when more than three stages are used. We therefore use three-stage networks throughout our experiments. A forward pass takes less than 1ms on a 1080Ti GPU. We can therefore say that the computation time of our multi-view regression model is almost negligible compared to the use of the 2D detector. 

\subsection{Datasets}

\miniTitle{Human3.6M. } Human3.6M is currently the largest multi-view human pose estimation dataset. The dataset includes around 3.6 million images collected from 15 actions performed by seven professional actors in a laboratory environment \cite{ionescu_pami2014}. The actions have been recorded by a four-view RGB camera system and camera parameters, including both intrinsic and extrinsic parameters, are available. Full-body 3D ground truth annotations are generated using a motion capture system. Following the standard evaluation protocol used in the literature, five subjects (S1, S5, S6, S7, S8) are used for training and two subjects (S9, S11) for testing \cite{chen_cvpr2017,pavlakos_cvpr2017,martinez_iccv2017}. Mean per joint position error (MPJPE) in millimeter is used as evaluation metric and test results are collected per action. 

\miniTitle{Multi-view OR.} The multi-view OR (MVOR) dataset is, to the best of our knowledge, the first multi-view pose estimation dataset that is generated 
from recordings in an uncontrolled environment. All activities in an operating room have been recorded for four days using a three-view camera system \cite{kadkhoda_wacv2017}.  We have selected every 1500 multi-view frames if there is at least one persons in one of the views. The dataset has been manually annotated to provide both 2D and 3D upper-body poses. 
The dataset includes around 700 multi-view frames and 1100 persons. The presence of multiple persons and clutter make this dataset much more challenging than Human3.6M as can be seen in Figure \ref{fig:mvFrame}. To report 2D body part localization on this dataset, we use the probability of correct keypoints (PCK) metric that is commonly used for evaluating multi-person pose estimation \cite{kadkhoda_wacv2017,cao_cvpr2017}. MPJPE is used to report 3D body part localization. 


\subsection{2D Detection Results}
\label{sec:2dDectRes}

\begin{table}[tb]
\centering
\setlength{\tabcolsep}{3pt} 
\renewcommand{\arraystretch}{0.9}
\begin{tabular}{l cc c c c c |c}
\toprule
Camera ID&Hip& Knee&	Foot& Shlder&	Elbow&	Wrist& Avg \\ 
\midrule

54138969& 16&	15&	14&	7&	11&	16&	13 \\
55011271& 13&	8&	10&	6&	8&	10&	9 \\
58860488& 15&	13&	12&	7&	11&	18&	13 \\
60457274& 16&	9&	10&	7&	10&	11&	11 \\
\hline
Avg & 15&	11&	11&	7&	10&	14&	11 \\
\bottomrule
\end{tabular}
\caption{2D MPJPE results of the 2D detector \cite{cao_cvpr2017} on the train set of Human3.6M. For each body part, the average MPJPE in pixels is computed across all actions in Human3.6M per camera.}
\label{tab:2dRes}
\end{table}

\begin{table}[tb!]
\centering
\setlength{\tabcolsep}{3pt} 
\renewcommand{\arraystretch}{0.9}
\begin{tabular}{l cc c c c |c}
\toprule
& Head &	Shlder&	Elbow&	Wrist& Hip&Avg \\ 
\midrule
Deep3DPS \cite{kadkhoda_wacv2017} & 93.4 & 77.0 & 71.5& 73.7 & 69.1 & 76.9 \\
\citenameref{cao_cvpr2017} &	92.8 & 90.1 & 75.6 & 75.9& 58.9 & 78.6 \\
\bottomrule
\end{tabular}
\caption{PCK results on MVOR. Body part detection results are reported for both Deep3DPS \cite{kadkhoda_wacv2017} and \citenameref{cao_cvpr2017} using the PCK metric. Note that Deep3DPS has been finetuned on another dataset captured in the same OR and relies on both color and depth images.}
\label{tab:2dResMVOR}
\end{table}

In this section, we evaluate the 2D detection model of \cite{cao_cvpr2017} on both datasets to assess its performance on such unseen data. In addition, we use the results on Human3.6M to model the characteristics of the 2D detector, which are required by our data generation model presented in Section~\ref{sec:trainDataGen}.  


In Table \ref{tab:2dRes}, we present the results of the single-view 2D pose detector \cite{cao_cvpr2017} on the Human3.6M train set. We should note that the detector has not seen any data from this dataset during training. We use MPJPE in pixel to compute body part localization errors. The results for each body parts are reported per camera. The results for head and neck localizations are not presented as the annotation for these body parts are different between Human3.6M and MS COCO that is used to train the detector. Note that the detector is applied on the whole image, i.e. no bounding box is provided, in contrast to previous work that relies either on ground truth  \cite{ionescu_pami2014,moreno_cvpr2017,martinez_iccv2017} or on person detectors \cite{tekin_cvpr2016} to obtain bounding boxes. In total, $3\%$ of the joints are not detected and the detector achieves the average MPJPE of 11 pixels. It is worth mentioning that the detector performs similarly on the test set. 
Table \ref{tab:2dResMVOR} presents the results of the 2D detector on the MVOR dataset. 
The model attains an average PCK of 78.9\% on this dataset. We have also reported the performance of Deep3DPS \cite{kadkhoda_wacv2017}, which is the state-of-the-art model on this dataset. In contrast to \cite{cao_cvpr2017}, which is trained on the RGB images of MS COCO, the Deep3DPS model uses both color and depth images and has been trained on MPI Pose and then finetuned on a single-view OR dataset. The 2D pose detector of \cite{cao_cvpr2017} outperforms Deep3DPS. These results show that the detector achieves fairly promising results on both datasets even without finetuning. Comparing the performance of the 2D detector on these two datasets also indicates that the MVOR dataset is much more complex, as the number of undetected joints is much higher ($21\%$ vs. $3\%$).  

For generating the training data, the evaluation results on the train set of Human3.6M, which are reported in Table \ref{tab:2dRes}, are used to set the parameters of the noise model. The train set from Human3.6M is chosen to avoid any overlap between train and test sets. The coefficient $\lambda$ in \eqref{eq:noiseConf} is set to two. As a result, $5\%$ of the joints will be labeled as undetected, which is on par with the percentage of undetected joints in Human3.6M.



\subsection{3D Localization Results}

\begin{table*}[!ht]
\centering
\setlength{\tabcolsep}{2.5pt} 
\renewcommand{\arraystretch}{0.9}
\resizebox{\textwidth}{!}{%
\begin{tabular}{lccccccccccccccc|c}
\toprule
Setting & Direc.&Discuss&Eat&Greet&Phone&Photo&Pose&Purch.&Sit&SitD&Smoke&Wait&Walk&WalkD&WalkT&Avg\\
\midrule
\citenameref{tekin_cvpr2016} & 102.4& 147.2& 88.8& 125.3& 118.0& 182.7& 112.4& 129.2& 138.9& 224.9& 118.4& 138.8& 126.3& 55.1& 65.8& 125.0 \\
\citenameref{chen_cvpr2017} & 89.9& 97.6& 89.9& 107.9& 107.3& 139.2& 93.6& 136.0& 133.1& 240.1& 106.6& 106.2& 87.0& 114.0& 90.5& 114.1 \\
\citenameref{pavlakos_cvpr2017} & 67.4 & 71.9& 66.7& 69.1& 72.0& 77.0& 65.0& 68.3& 83.7& 96.5& 71.7& 65.8& 74.9& 59.1& 63.2& 71.9 \\
\citenameref{martinez_iccv2017} &51.8& 56.2& 58.1& 59.0& 69.5& 78.4& 55.2& 58.1&  74.0 & 94.6& 62.3& 59.1& 65.1& 49.5& 52.4 & 62.9\\

\hdashline
$<$SV, \citenameref{newell_eccv2016}$>$ &53.4&58.6&62.1&63.2&86.2&83.3&56&58.1&81.2&101.2&68.4&64.1&67.4&51&54.2&67.2\\
$<$SV, \citenameref{cao_cvpr2017}$>$ & 69.5 & 75.5 & 67.6 & 76.8 & 84.6 & 94.9 & 69.8 & 68.4 & 92.2 & 113.7 & 77.1 & 75.1 & 77.2 & 59.0 &  64.2 & 77.7 \\

\hdashline
$<$SV, GT$>$ & 94.2 & 113.7 & 96.9 & 106.5 & 119.8 & 127.6 & 86.5 & 149.9 & 145.6 & 222.3 & 113.5 & 111.2 & 120.9 & 92.8 & 92.4 & 119.6 \\
$<$SV, Noisy GT$>$ & 69.7 & 78.8 & 69.8 & 77.5 & 84.4 & 97.6 & 64.9 & 86.5 & 103.3 & 125.8 & 81.8 & 80.4 & 83.3 & 59.9 & 62.6 & 81.8 \\

\hdashline
\citenameref{pavlakos_cvpr2017_2} & 41.2 & 49.27 & 42.8 & 43.5 & 55.6 & 46.9 & 40.3 & 63.7 & 97.6 & 119.9 & 52.1 & 42.7 & 41.8 & 51.9 &  39.4 & 56.9 \\

$<$MV, \citenameref{cao_cvpr2017}$>$ & 39.4&46.9&41.0&42.7&53.6&54.8&41.4&50.0&59.9&78.8&49.8&46.2&51.1&40.5&41.0&49.1 \\
$<$MV, GT$>$ & 92.1&105.8&110.1&94.0&128.2&117.0&77.0&152.2&152.0&227.5&122.9&104.3&125.1&88.7&80.9&118.5 \\
$<$MV, Noisy GT$>$ & 47.1&60.5&48.7&53.5&63.5&71.1&48.7&57.8&72.2&81.7&59.0&55.9&60.6&43.4&44.3&57.9\\
\bottomrule
\end{tabular} 
}
\caption{3D MPJPE in millimeter on Human3.6M. Quantitative results of our approach with different configurations are reported and compared with the state-of-the-art. SV indicates that our model is trained on single-view input and MV indicates multi-view input. Our models are trained on 2D poses obtained from: \cite{newell_eccv2016}'s model, \cite{cao_cvpr2017}'s model, 2D projections of 3D ground truths (GT) and our data generation method described in Section \ref{sec:catDet} (Noisy GT). $<$M, P$>$ denotes our model settings where M is either SV or MV and P indicates the type of poses that are used during training.}
\label{tab:h36m}
\end{table*}

\miniTitle{Human3.6M.} 
As Human3.6M is a fairly new dataset and state-of-the-art results are mainly reported using single-view models, we compare our approach with recent state-of-the-art single- and multi-view models for 3D pose estimation on Human3.6M. For the sake of comparison, we have therefore trained a variant of our proposed regression function that relies solely on single-view input. Table \ref{tab:h36m} reports evaluation results of our approach with different configurations. Models that are relying on single-view input are denoted by SV and multi-view ones by MV. These models are trained either on ground truth (GT) 2D poses, Noisy GT 2D poses as described in Section \ref{sec:trainDataGen} or on 2D detections provided by either \cite{newell_eccv2016} or \cite{cao_cvpr2017} for comparison. Even though Human3.6M is a single-person dataset, note that in \cite{tekin_cvpr2016,pavlakos_cvpr2017} the input images are cropped using bounding boxes around the persons and that the 2D pose detector models of \cite{newell_eccv2016} and \cite{wei_cvpr2016} used in \cite{chen_cvpr2017} and \cite{martinez_iccv2017} are applied on bounding boxes around the persons obtained from ground truth. 


Our single-view 3D pose regression model trained on 2D detection provided by \cite{newell_eccv2016} achieves the average localization error of $67.2$ mm. We should note that our results for this model improve slightly over the results  reported by \cite{martinez_iccv2017} on the same experimental setup ($67.5$), where the same 2D pose detector trained on MPII Pose is used without any finetuning on Human3.6M.   
\cite{martinez_iccv2017} showed that the results can be improved by finetuning the model on Human3.6M (62.9 vs. 67.5), which is in line with the results reported in \cite{chen_cvpr2017}. However, in order to easily generalize to new environments, we do not finetune 2D pose detectors as this would require annotated data. Except the model $<$SV, \cite{newell_eccv2016}$>$, which uses the same 2D pose detector during both training and testing for the sake of fair comparison with \cite{martinez_iccv2017}, all our models have used 2D detections provided by \cite{cao_cvpr2017} during testing\footnote{Please note that at test time 2D poses are detected using \cite{cao_cvpr2017} even in case of models trained on GT poses, which is different from \cite{martinez_iccv2017}.}. We should note that even though our single-view 3D regression model trained on the 2D detections provided by \cite{newell_eccv2016} performs better than other variants of our single-view model, we decide to use the model of \cite{cao_cvpr2017} instead, as it is not restricted to bounding boxes and allows us to detect and estimate 2D body poses in {\it multi-person scenarios}, e.g. the MVOR dataset.

The evaluation results show that our single-view model trained on ground truth 2D poses and the model of \cite{chen_cvpr2017} perform similarly. This indicates that our regression function that is trained on perfect GT data will eventually work similarly to the lookup table used in \cite{chen_cvpr2017}. One can therefore conclude that if perfect 2D detections are obtained, a 2D-to-3D regression function or a lookup table would work similarly. But, the 2D detections are not perfect in practice. Therefore, by incorporating detection noise during training as described in Section \ref{sec:trainDataGen}, we have constructed a model $<$SV, Noisy GT$>$ that could cope better with noisy detection (81.8 vs. 119.6). We observe that if we train the model on 2D detections from the same 2D detector used during testing, i.e. \cite{cao_cvpr2017}, average MPJPE is improved by only four millimeters. These results indicate that our data generation model presented in Section \ref{sec:trainDataGen} has properly incorporated the detector's characteristics and our approach generalizes well to test data.

We have also presented the evaluation results of our multi-view regression function in Table \ref{tab:h36m}. Training the model $<$MV, \cite{cao_cvpr2017}$>$ on 2D pose detections by the same detector model as the one used at test time achieves the average MPJPE of 49 millimeters, which outperforms \cite{pavlakos_cvpr2017_2}. This is the lower limit for MPJPE on Human3.6M, which can be obtained by our MV regression model using this single-view pose detector. During our experiments, we observe that even though our multi-view regression models have generally converged to lower training losses compared to single-view ones, both single-view and multi-view models trained on ground truth poses achieve similar performance (119.6 vs. 118.5). We believe that as the multi-view model is only trained on perfect ground truth 2D poses, it always expects the exact projections of a 3D pose in all views. But, since the 2D pose detector provides noisy detections, this is not always possible at test time. The last row shows the results of our multi-view regression model trained using 2D poses generated from 3D ground truth by incorporating the 2D detector's characteristics. We should note that even without finetuning the detector on Human3.6M this model performs similarly to \cite{pavlakos_cvpr2017_2}, which has been trained on Human3.6M.  This model also reduces the error by more than $50\%$ compared to the same model trained on ground truth data only. Furthermore, the model has also improved the localization results by $\sim30\%$ compared to the single-view model $<$SV, Noisy GT$>$ indicating that this model has properly incorporated 2D body part locations across all views to regress for their 3D positions. These results also confirm our hypothesis that incorporating the characteristics of the detector during training enables developing models that are robust to the inaccuracies and failures of the detector at test time. 

\begin{table*}[!htb]
\centering
\setlength{\tabcolsep}{3pt} 
\renewcommand{\arraystretch}{0.9}
\begin{tabular}{lcccccc}
\toprule
\multirow{2}{*}{\begin{tabular}[c]{@{}l@{}}Part\\ name\end{tabular}} & \multicolumn{2}{c}{One view} & \multicolumn{2}{c}{Two views} & \multicolumn{2}{c}{Three views} \\ 
\cmidrule(lr){2-3} \cmidrule(lr){4-5} \cmidrule(lr){6-7}
& MVDeep3DPS & Ours & MVDeep3DPS & Ours & MVDeep3DPS & Ours \\
\midrule

Shoulder &19&13&15&8&10&5 \\
Hip &27&20&23&15&17&11 \\
Elbow &27&25&23&19&16&12 \\ 
Wrist & 32&34&25&28&18&16 \\
\hdashline
Average& 26&23&22&18&15&11 \\


\bottomrule
\end{tabular}
\caption{MPJPE in centimeter on MVOR. Quantitative results of our approach are  compared with MVDeep3DPS \cite{kadkhoda_wacv2017}. We follow the same convention as in MVDeep3DPS and report the results per number of supporting viewpoints for the same set of body parts.}
\label{table:mvctRes}
\end{table*}

\miniTitle{Multi-view OR.}
In order to assess the ability of our approach to generalize to new multi-view environments, we evaluate the performance of our approach on the multi-view OR dataset. We use the 3D poses from Human3.6M, the camera calibration parameters of MVOR and the data generation model described in Section \ref{sec:catDet} to train a multi-view 3D regression model. The evaluation results of this model on MVOR are presented in Table \ref{table:mvctRes}. We use 3D MPJPE in centimeter as evaluation metric. Following the convention in MVDeep3DPS \cite{kadkhoda_wacv2017}, MPJPE is computed for the same set of body parts and is reported per number of supporting views. Our model has achieved the average MJPJPE of 17 cm on this dataset. The results show a significant improvement in the localization of the body parts as the number of supporting view increases. The average MPJPE is improved by 12 cm for persons who are detected in three views compared to those who are only detected in one view. This clearly indicates the benefit of observing an environment from multiple complementary views and the ability of our regression model to leverage such data for predicting 3D body poses even when some body parts are invisible. 

Table \ref{table:mvctRes} also compares the performance of our model with the MVDeep3DPS model \cite{kadkhoda_wacv2017}. We should note that MVDeep3DPS requires both color and depth images in contrast to our approach that relies solely on color images. Our approach, which only uses Human3.6M data, improves the results over MVDeep3DPS, even though MVDeep3DPS is trained on an annotated dataset recorded in the same OR as the one used to capture MVOR. This evaluation results demonstrate that our approach can exploit existing datasets to easily generalize to new multi-view setups without any need for new annotations.  

\begin{figure*}[!htb]
\centering
\setlength{\tabcolsep}{3pt}
\renewcommand{\arraystretch}{1.5}
\begin{tabular}{c c c c r}
\includegraphics[width=0.18\linewidth]{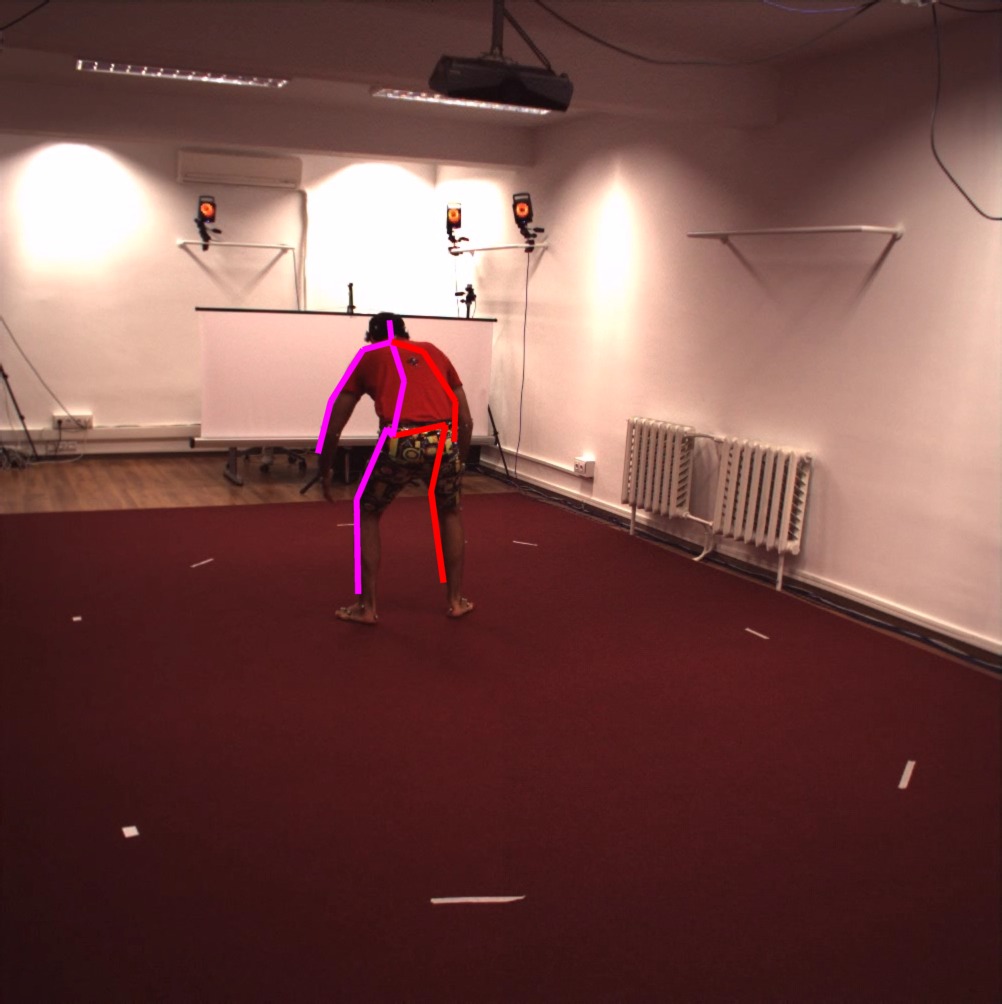} &
\includegraphics[width=0.18\linewidth]{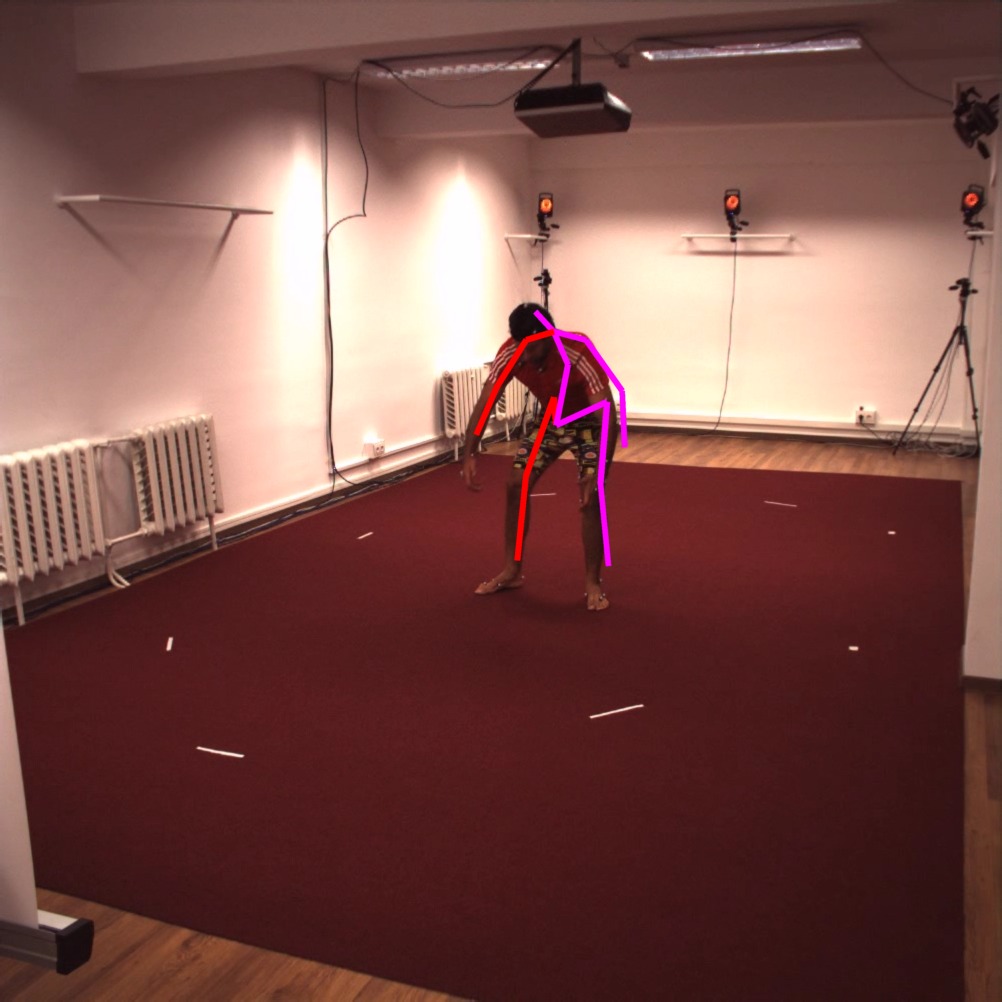} &
\includegraphics[width=0.18\linewidth]{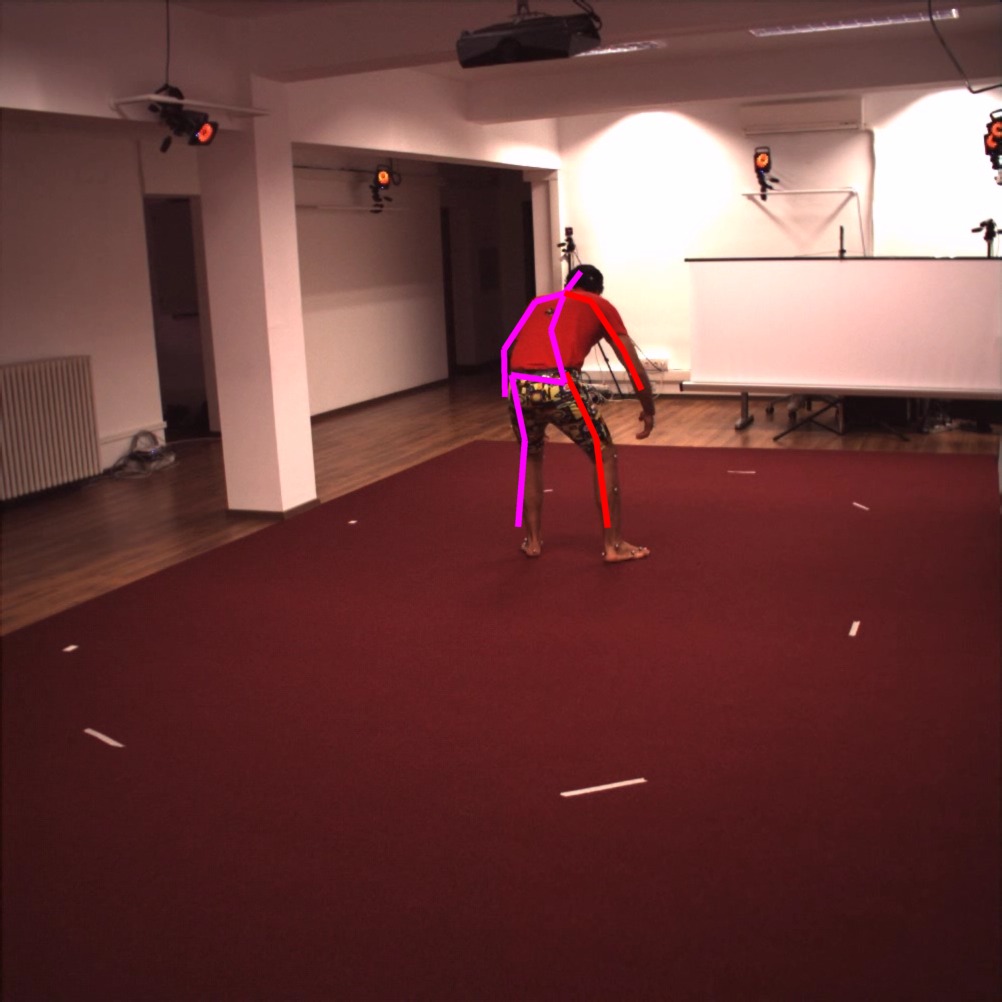} &
\includegraphics[width=0.18\linewidth]{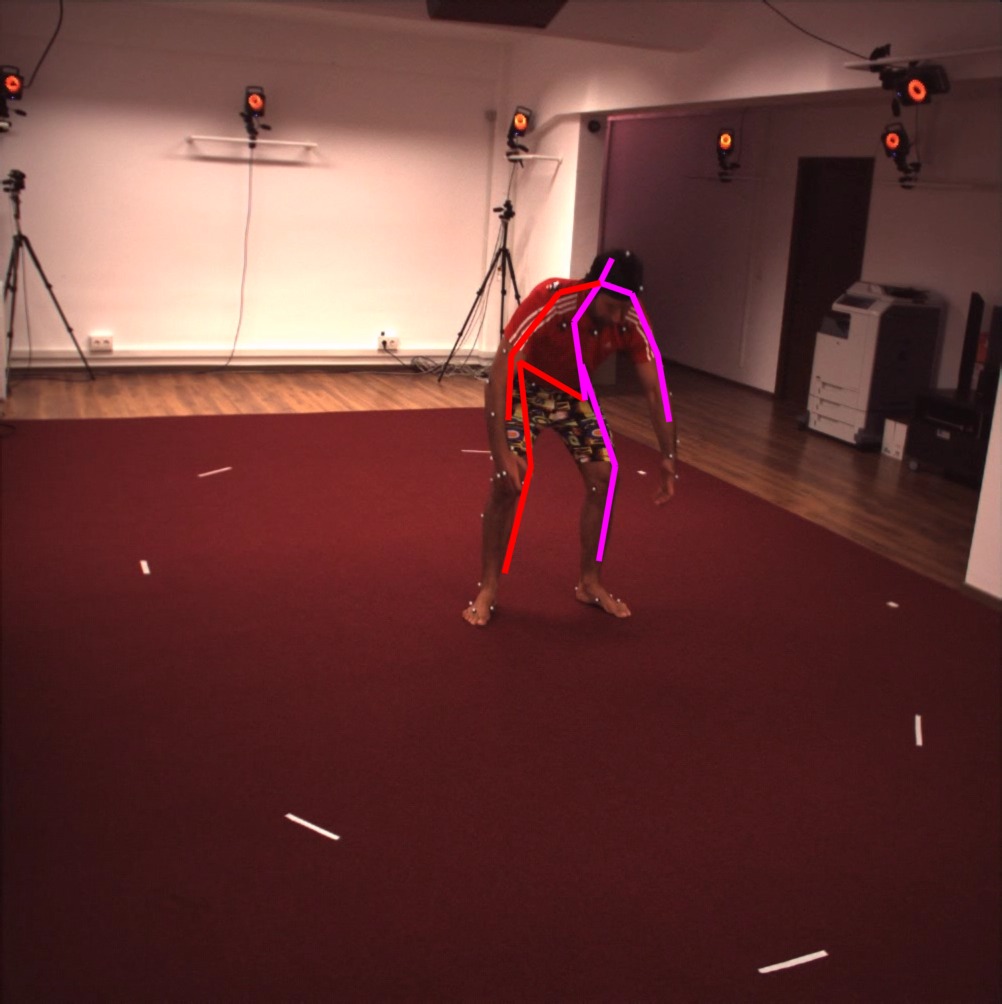} & 
\includegraphics[width=0.18\linewidth]{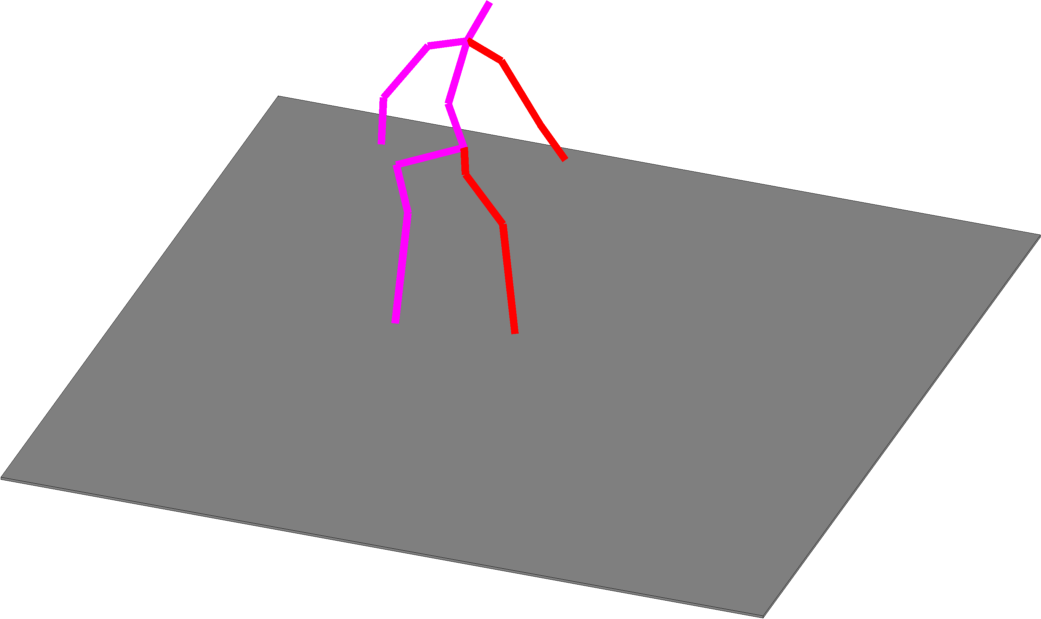} \\

\includegraphics[width=0.18\linewidth]{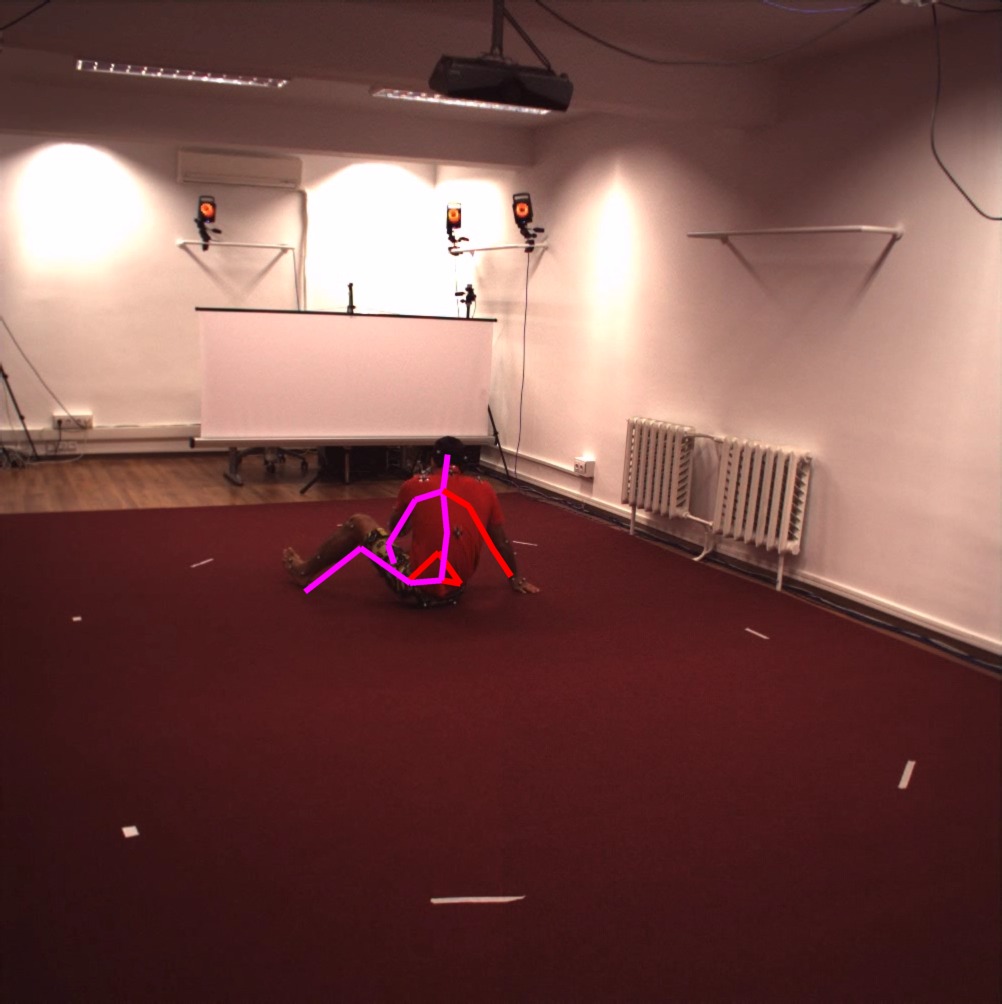} &
\includegraphics[width=0.18\linewidth]{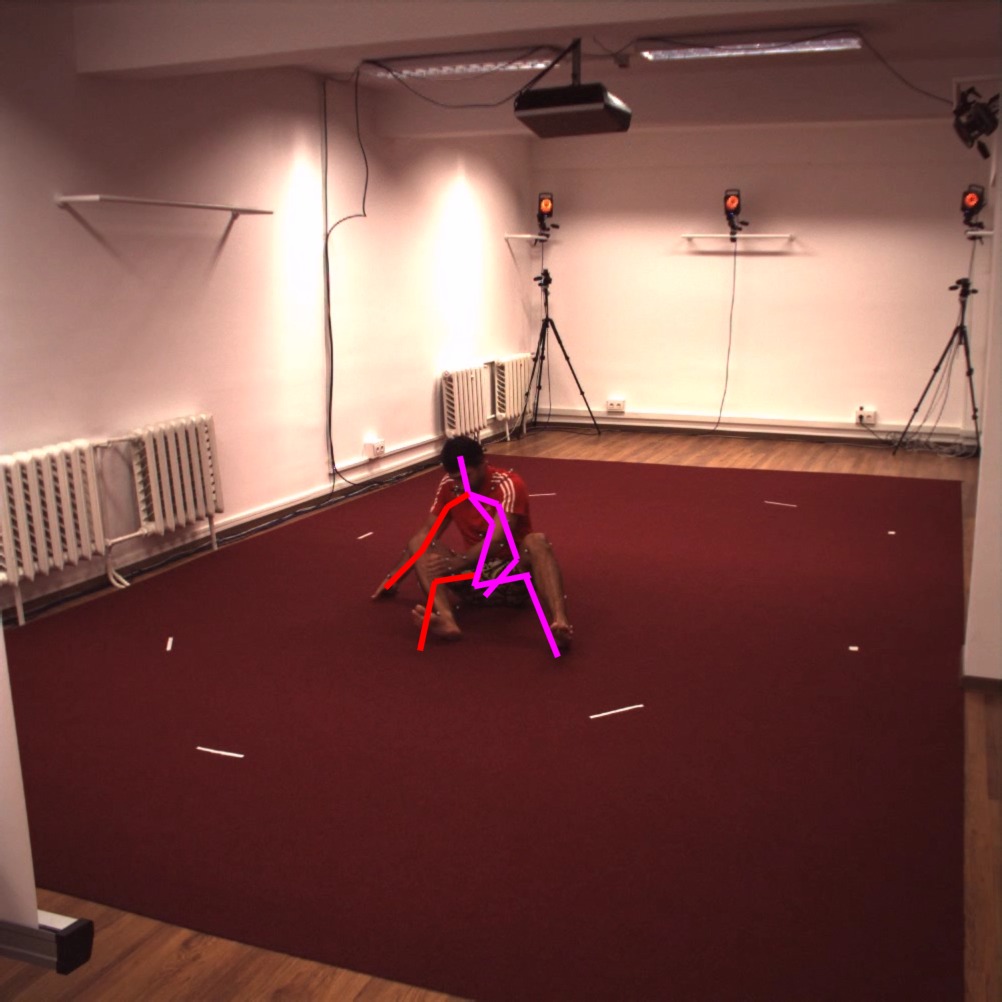} &
\includegraphics[width=0.18\linewidth]{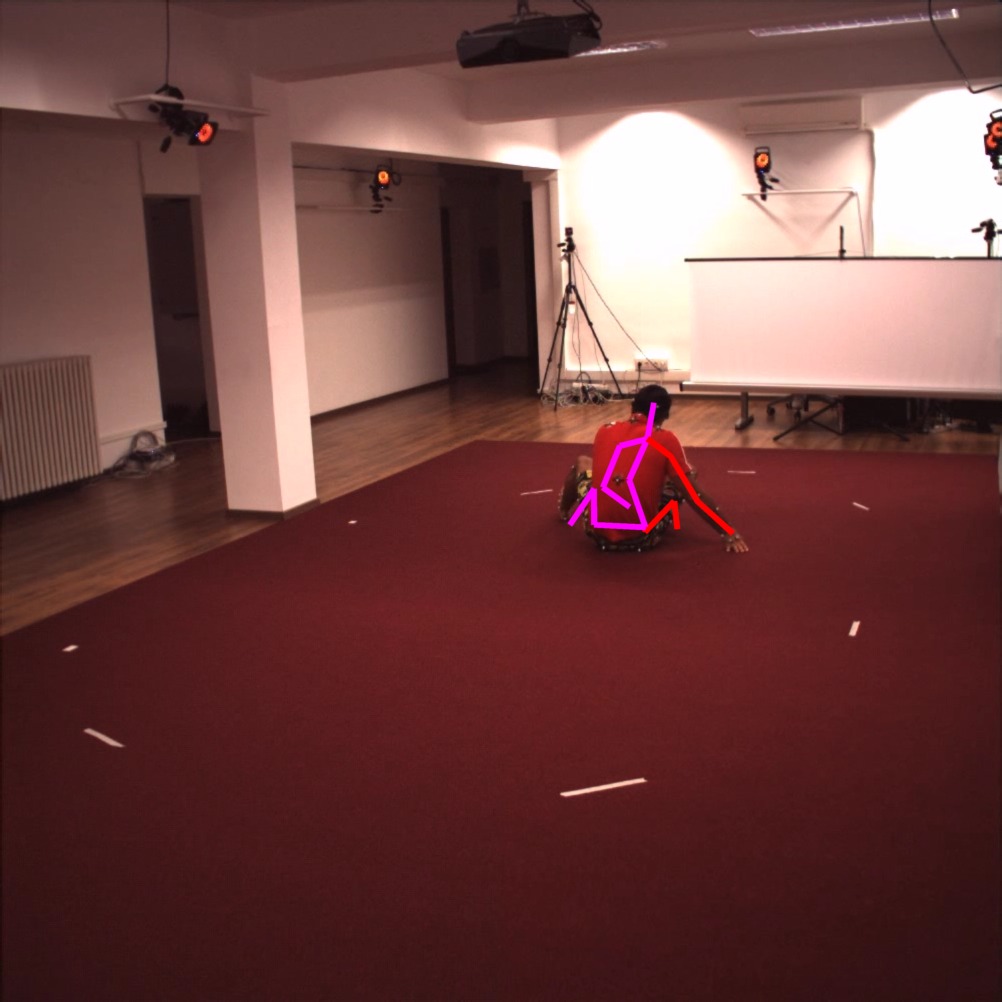} &
\includegraphics[width=0.18\linewidth]{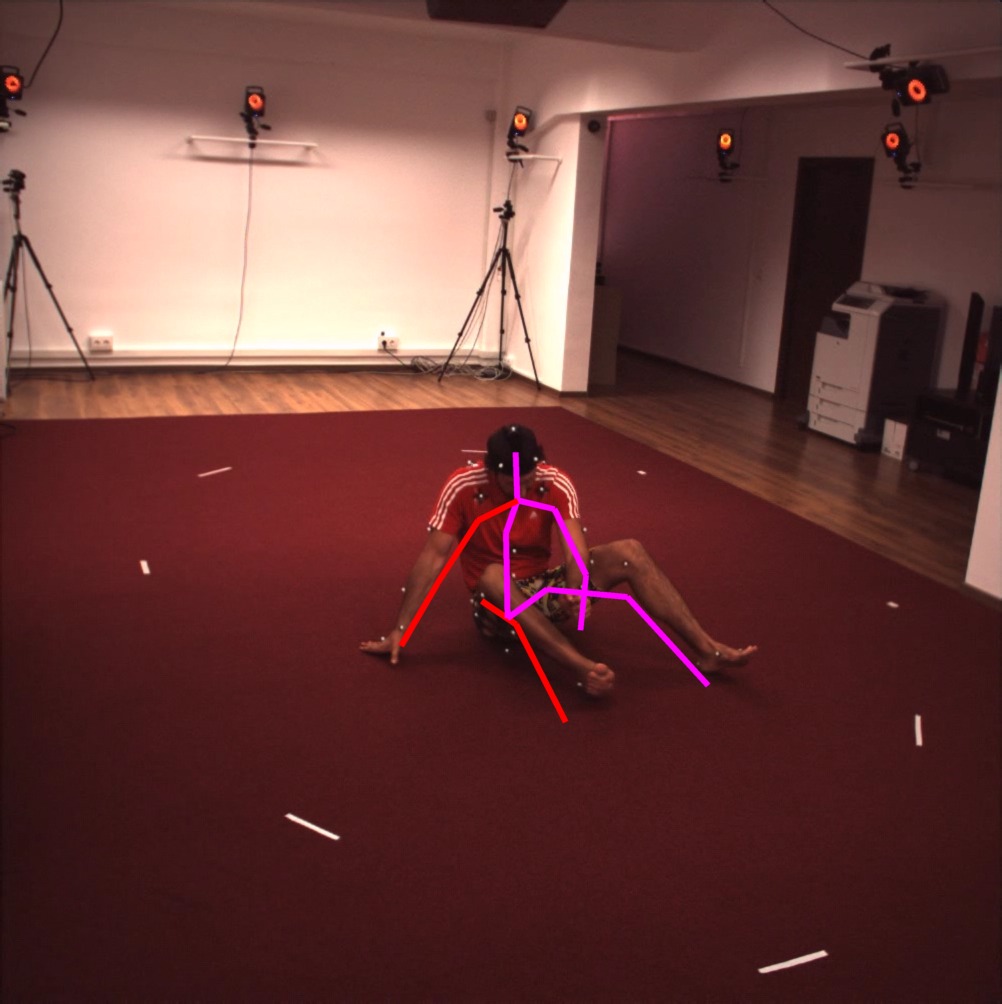} &  
\includegraphics[width=0.18\linewidth]{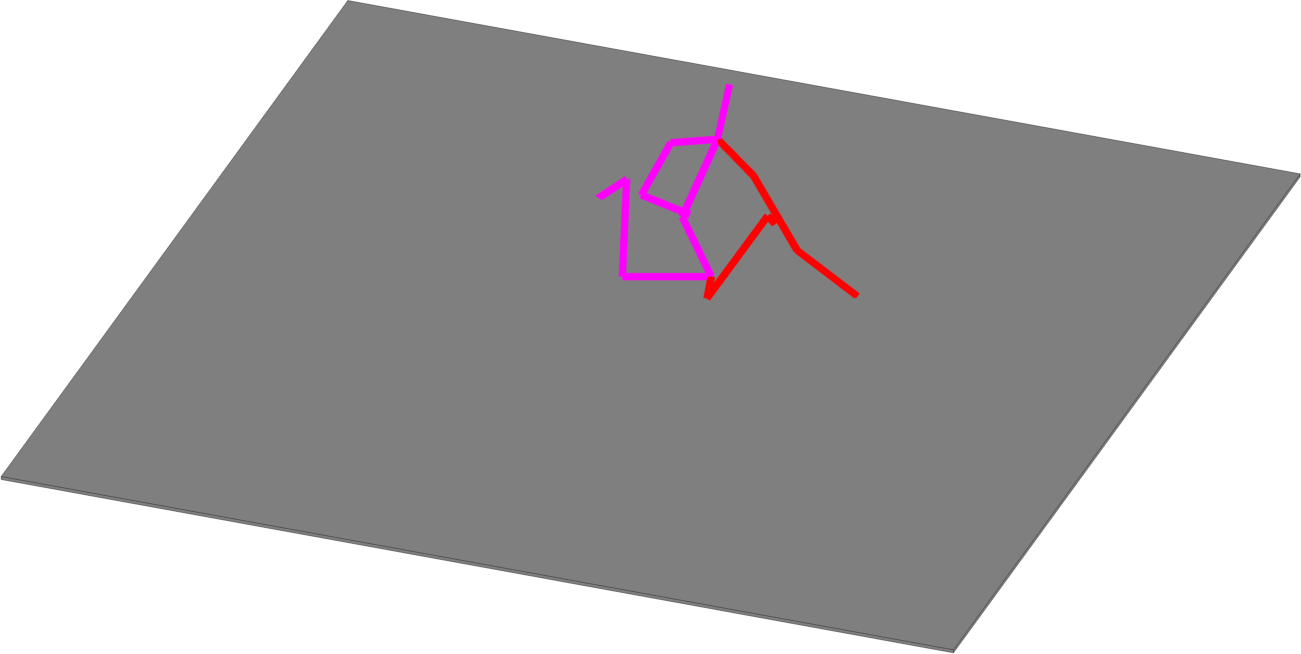} \\

\includegraphics[width=0.18\linewidth]{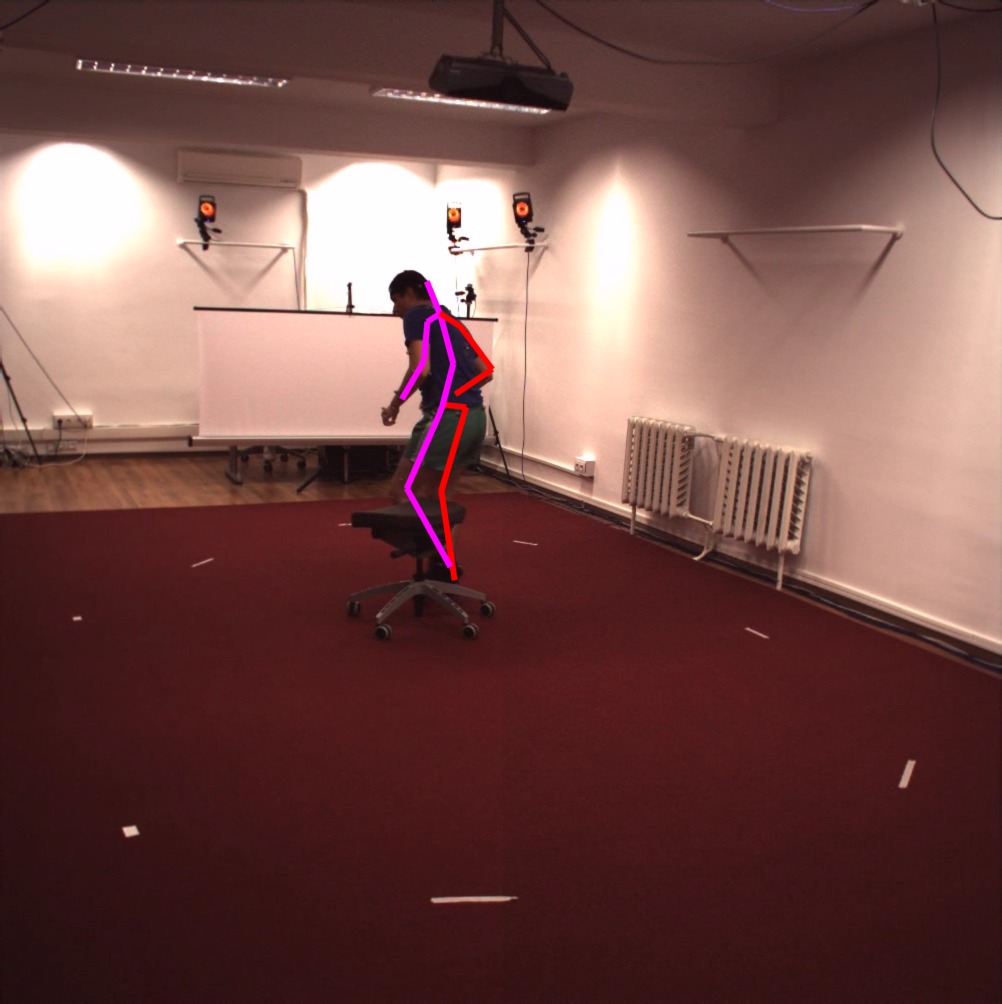} &
\includegraphics[width=0.18\linewidth]{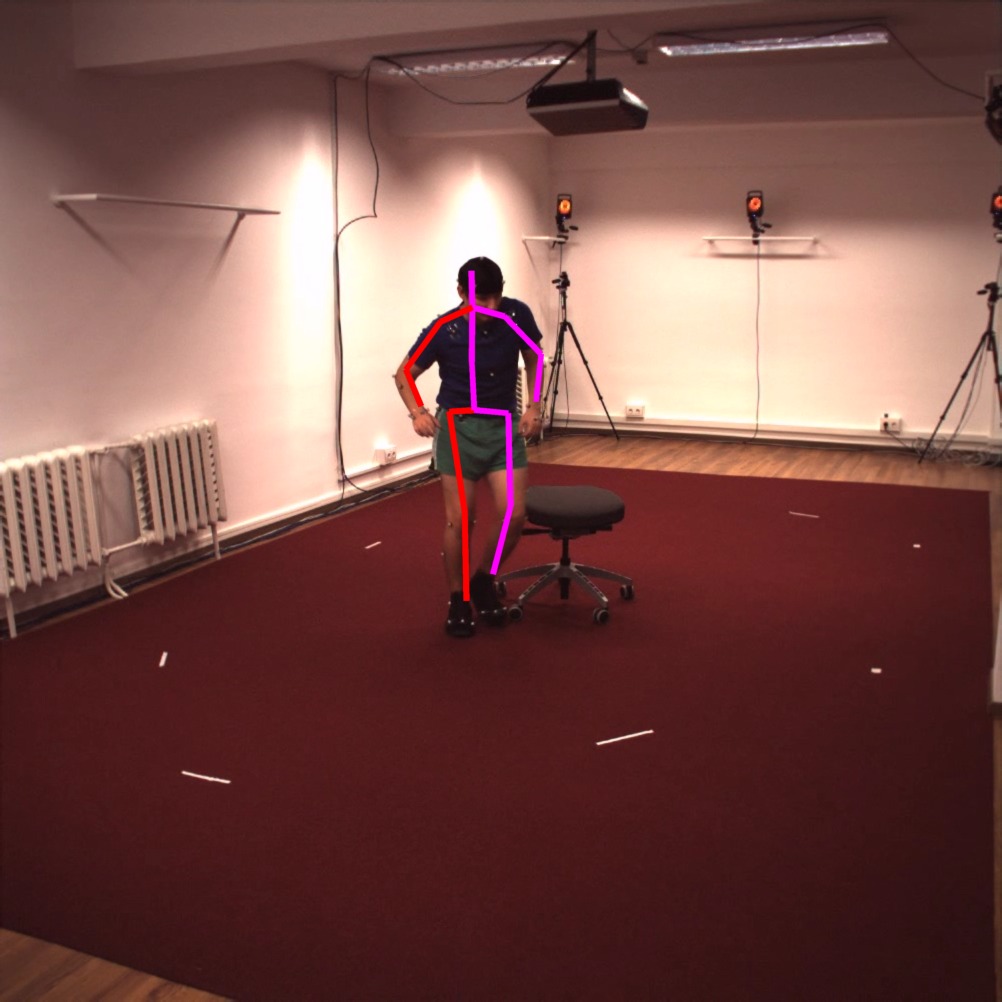} &
\includegraphics[width=0.18\linewidth]{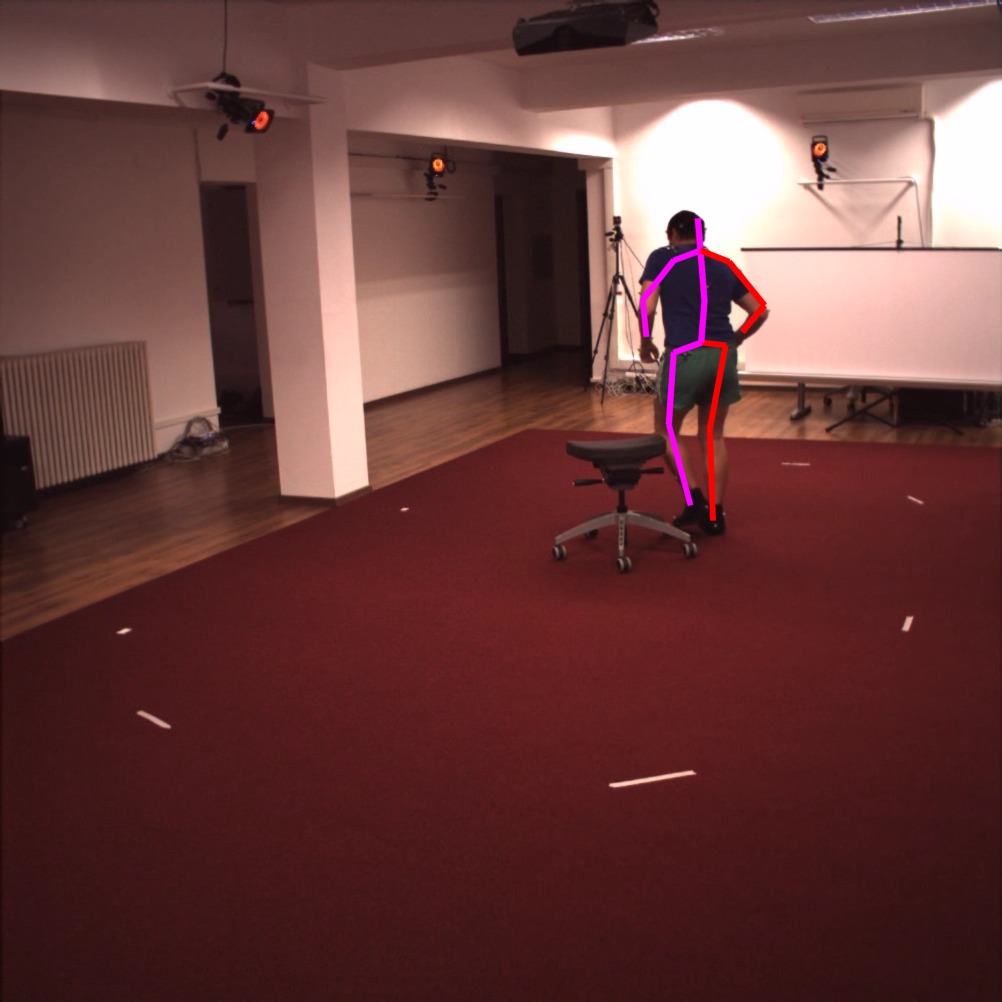} &
\includegraphics[width=0.18\linewidth]{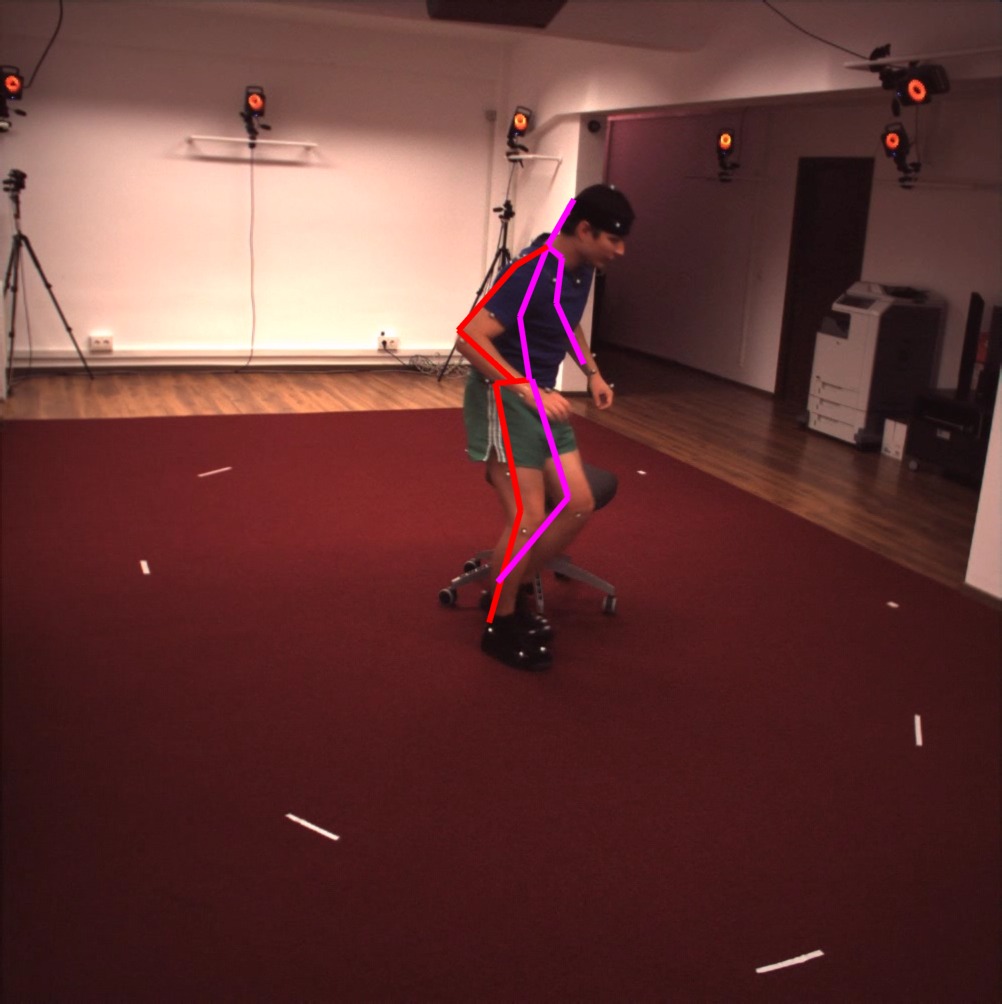} &  
\includegraphics[width=0.18\linewidth]{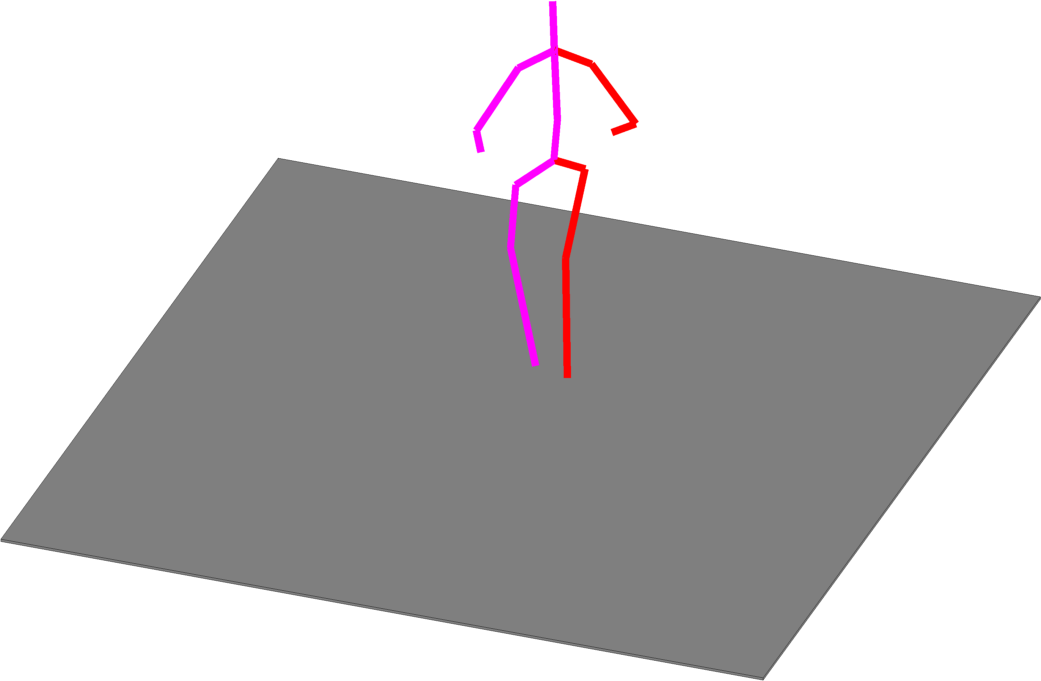} \\

\includegraphics[width=0.18\linewidth]{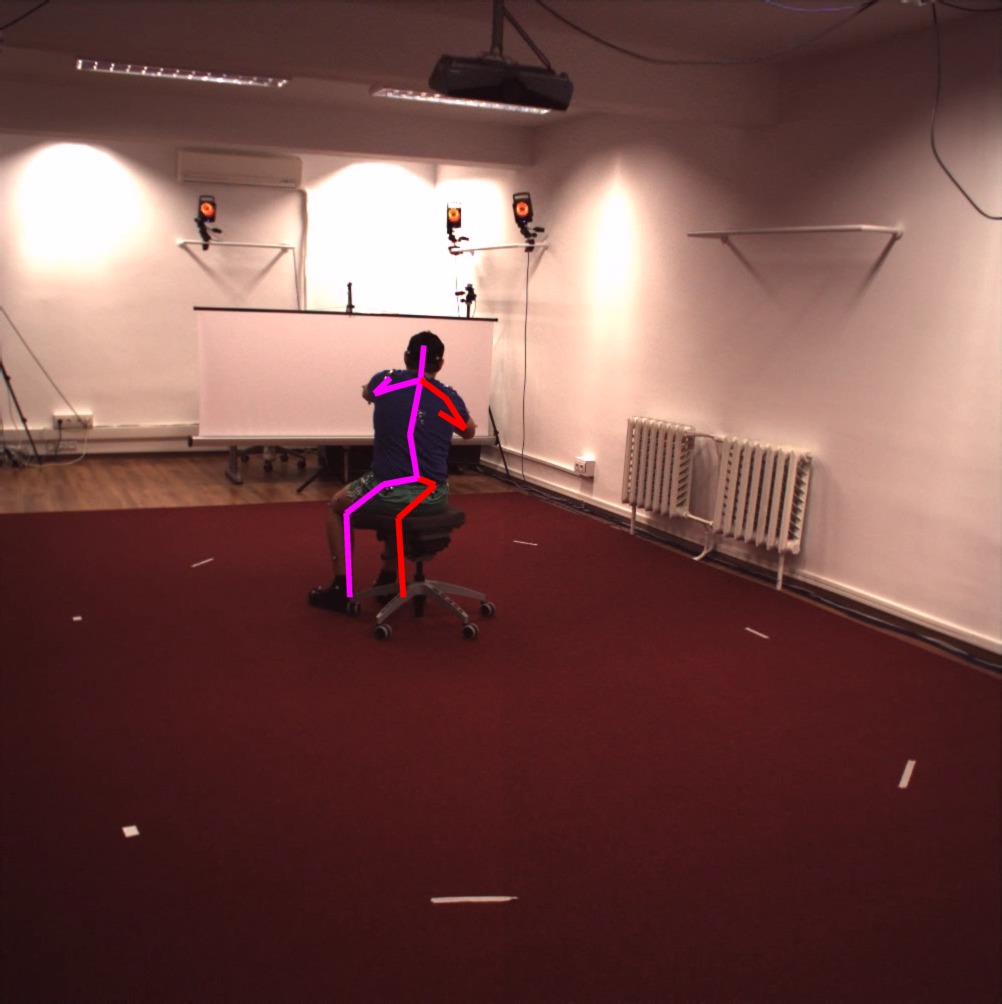} &
\includegraphics[width=0.18\linewidth]{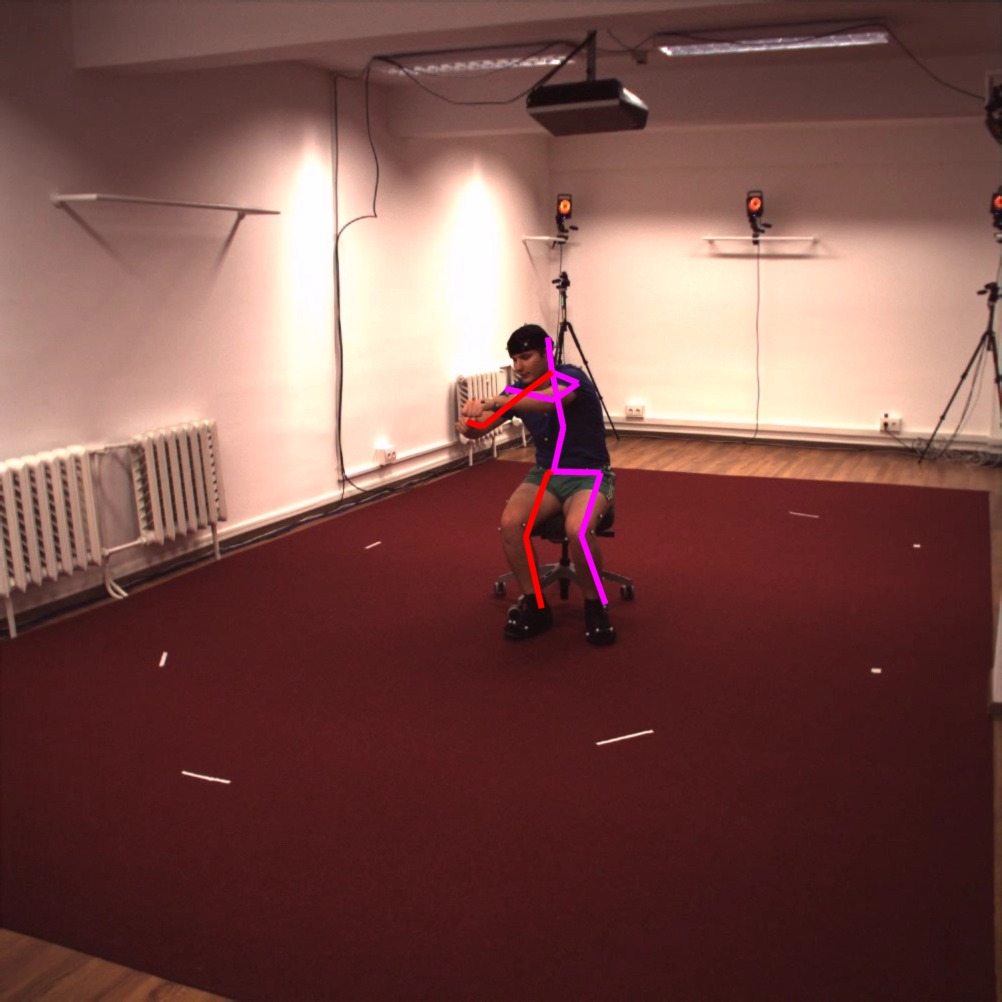} &
\includegraphics[width=0.18\linewidth]{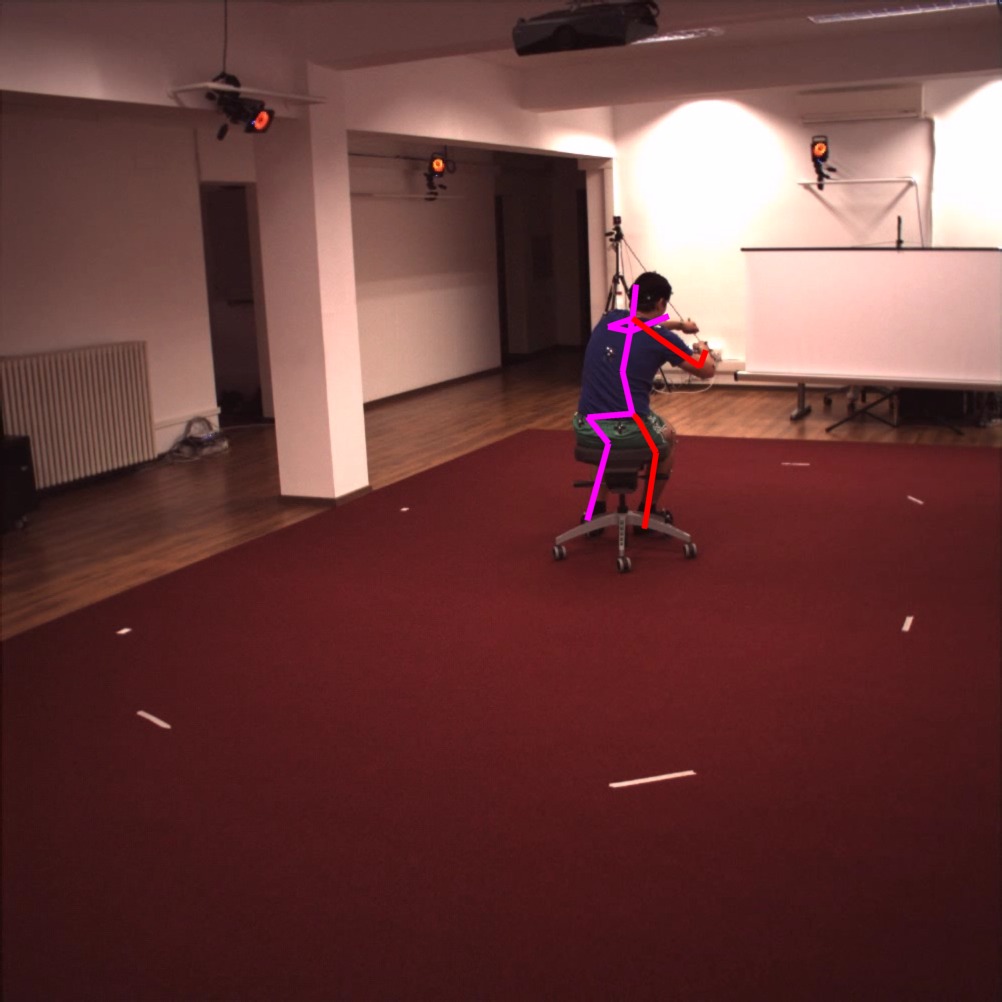} &
\includegraphics[width=0.18\linewidth]{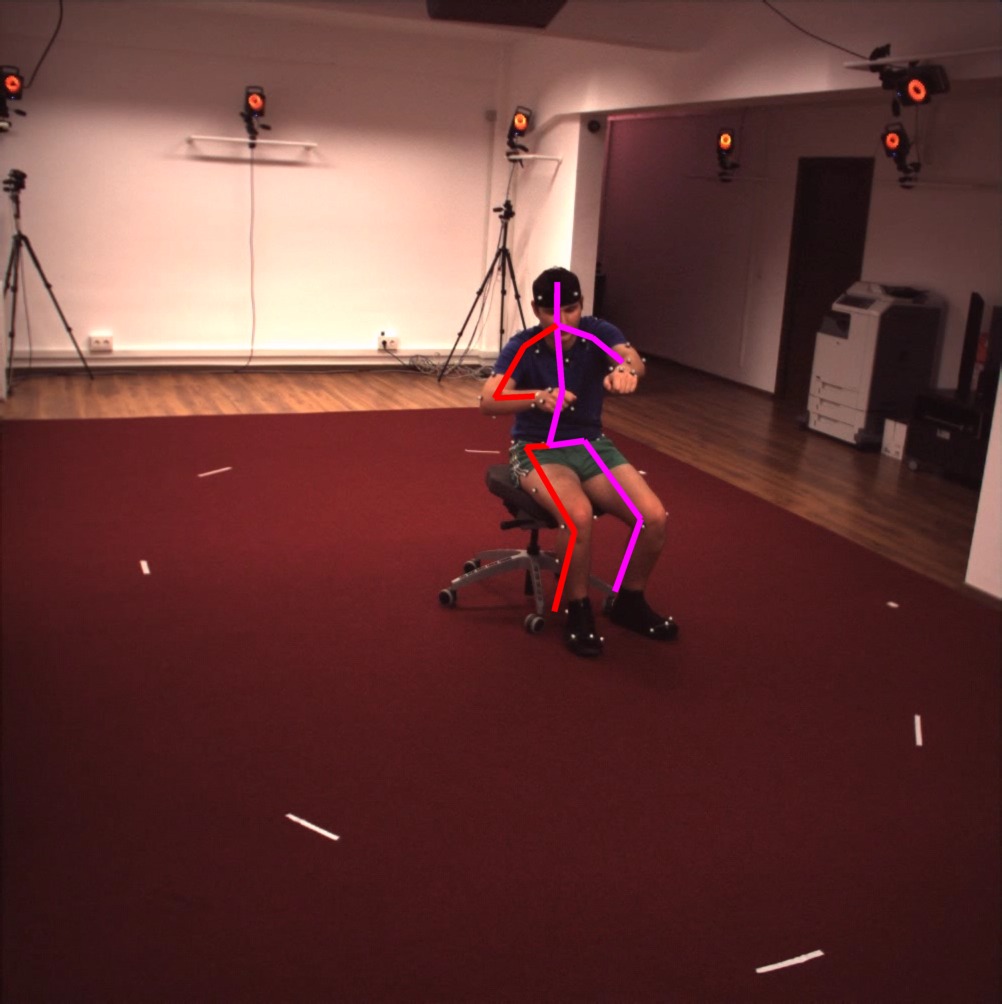} &  
\includegraphics[width=0.18\linewidth]{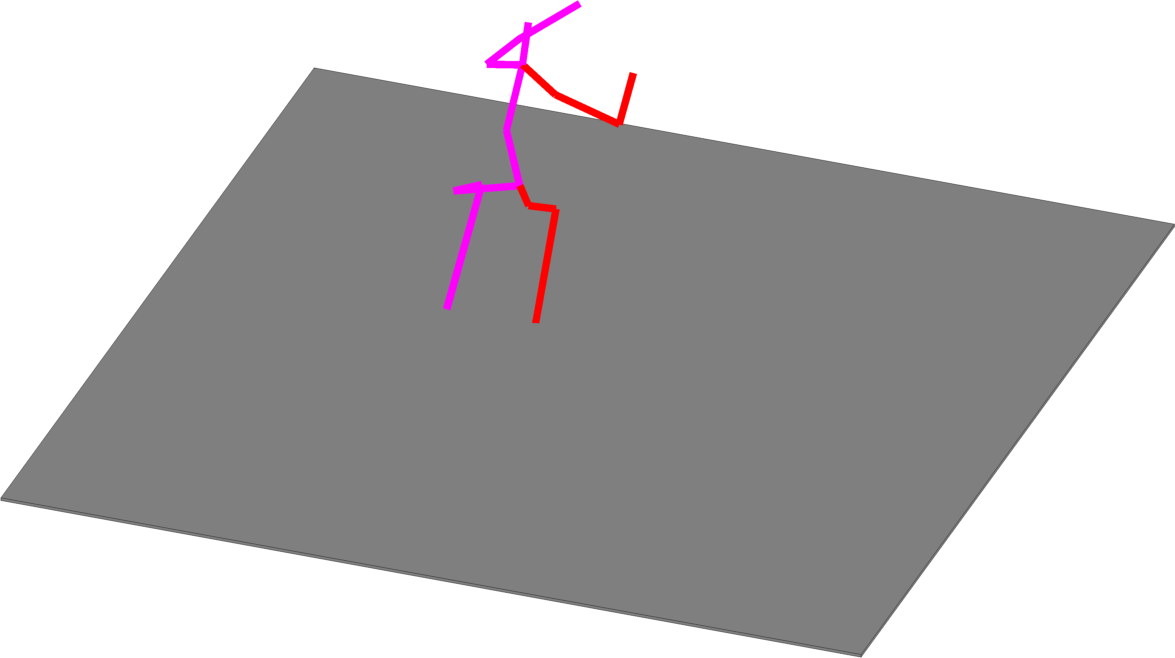} \\

\includegraphics[width=0.18\linewidth]{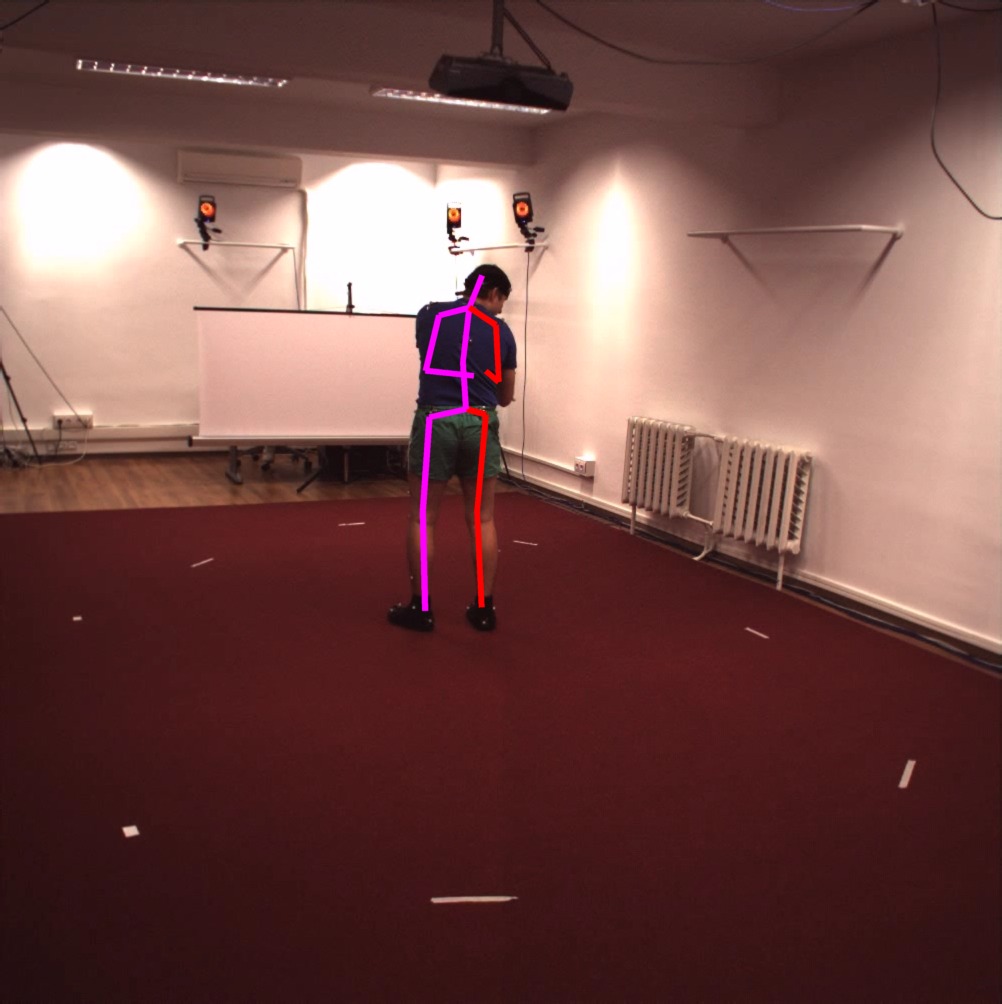} &
\includegraphics[width=0.18\linewidth]{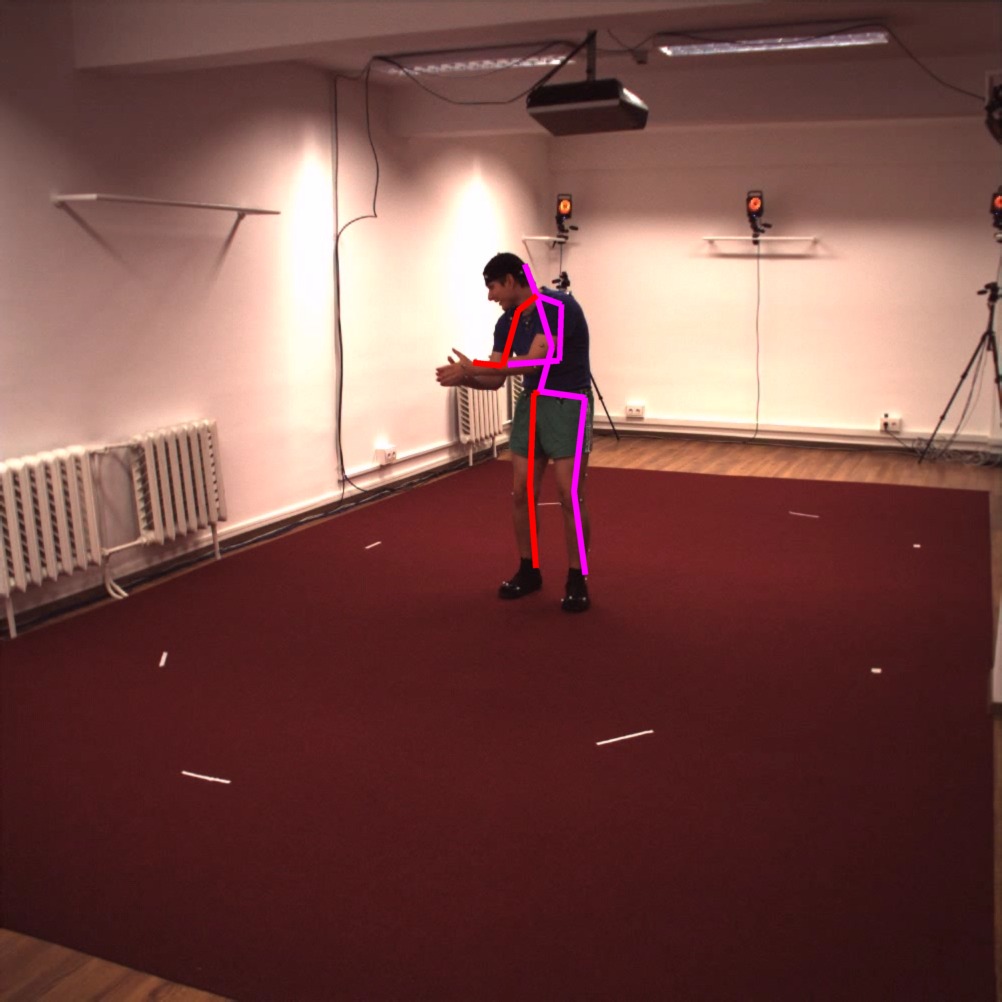} &
\includegraphics[width=0.18\linewidth]{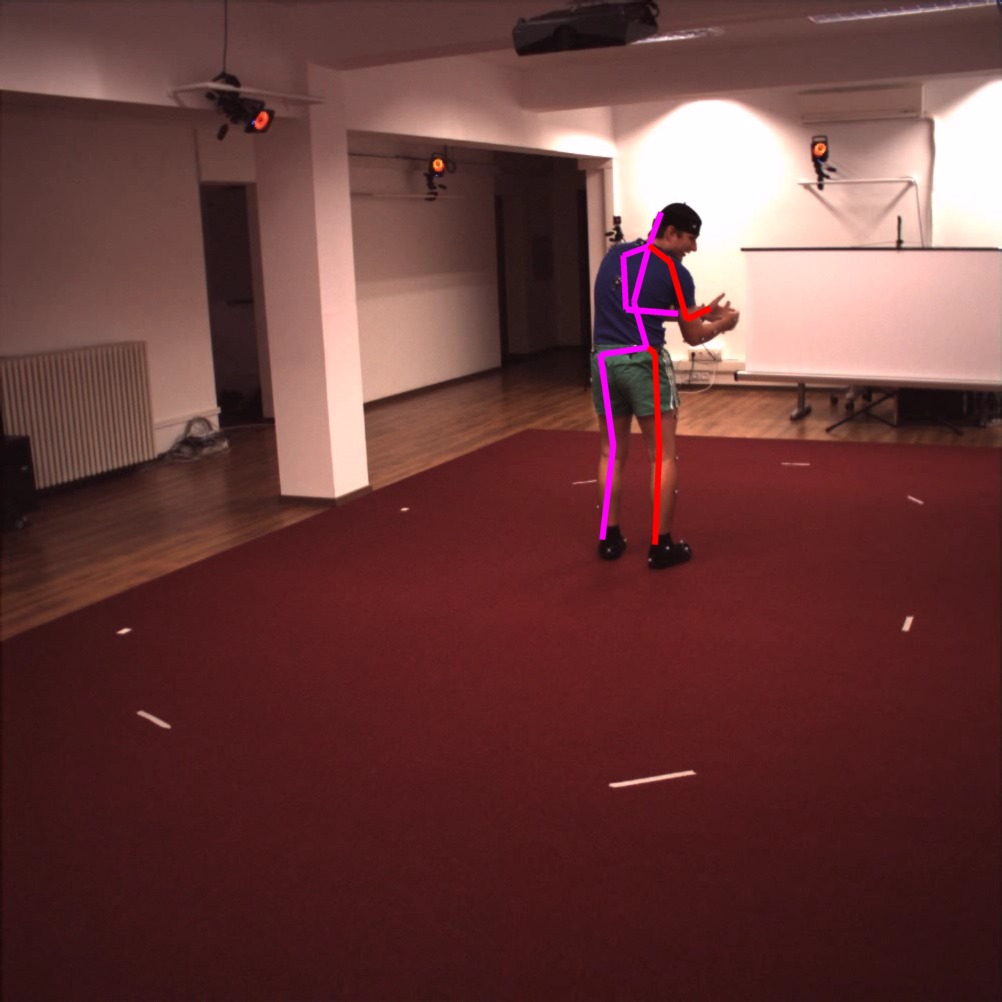} &
\includegraphics[width=0.18\linewidth]{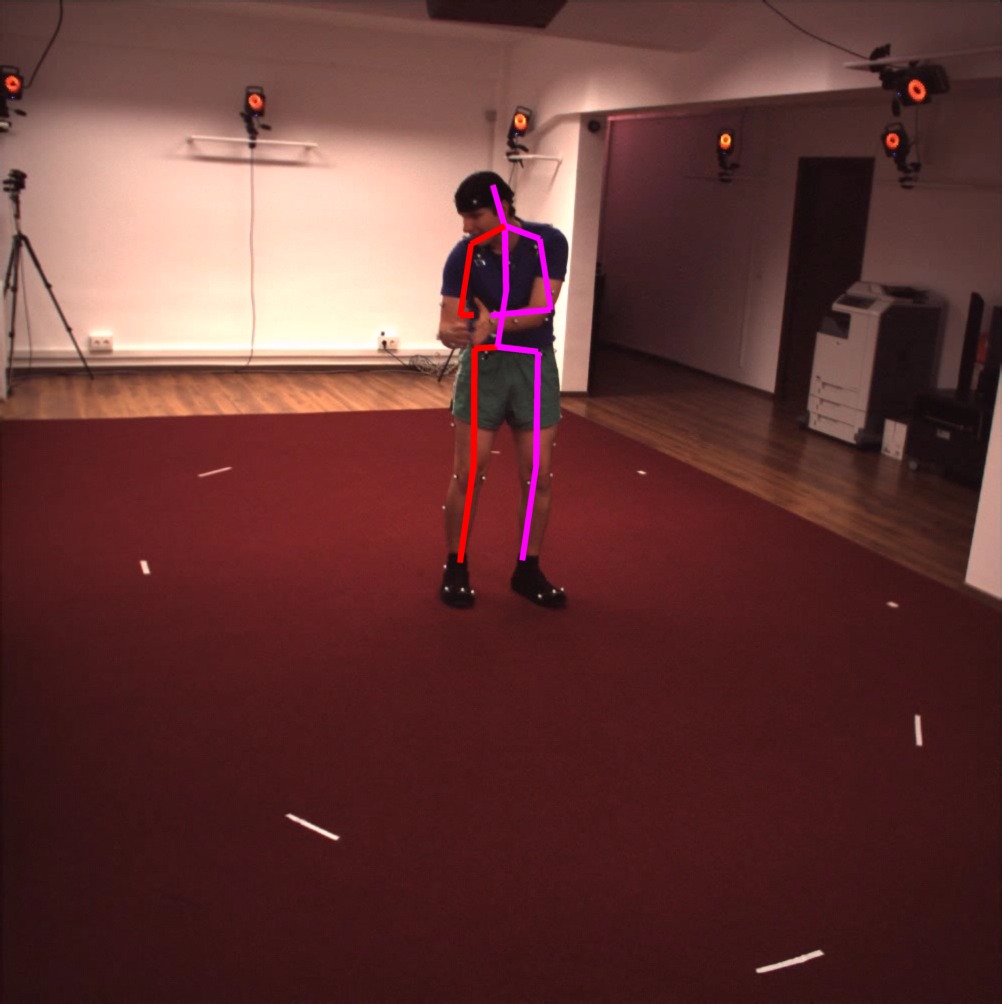} &  
\includegraphics[width=0.18\linewidth]{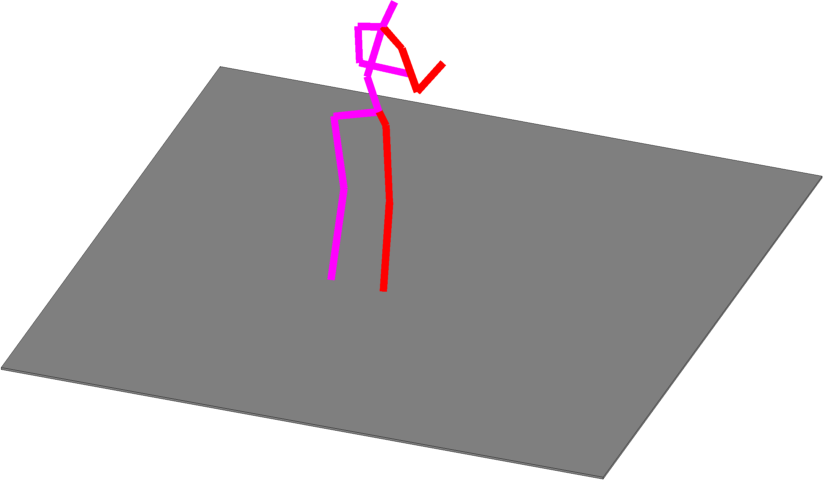} 

\end{tabular}

\caption{Qualitative results on the Human3.6M dataset. The last column shows the 3D poses and the other columns show the corresponding multi-view frames where the projected 2D poses are superimposed. The body parts from the right side of the body are drawn in red. (Best seen in color) }
\label{fig:h36mQual}
\end{figure*}

\begin{figure*}[!htb]
\centering
\setlength{\tabcolsep}{3pt}
\renewcommand{\arraystretch}{1.5}
\begin{tabular}{c c c c r}
\includegraphics[width=0.24\linewidth]{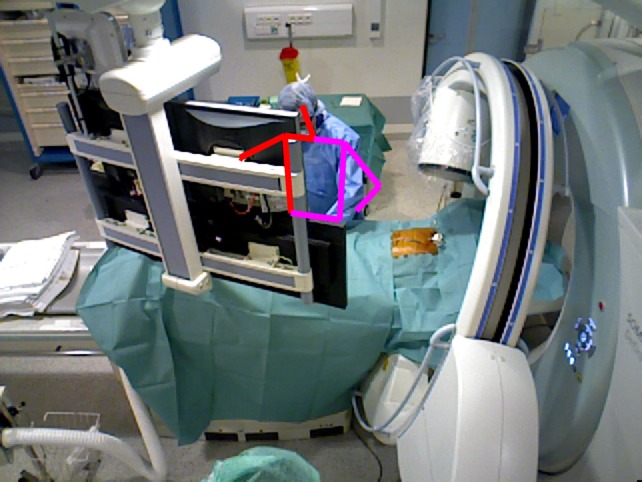} &
\includegraphics[width=0.24\linewidth]{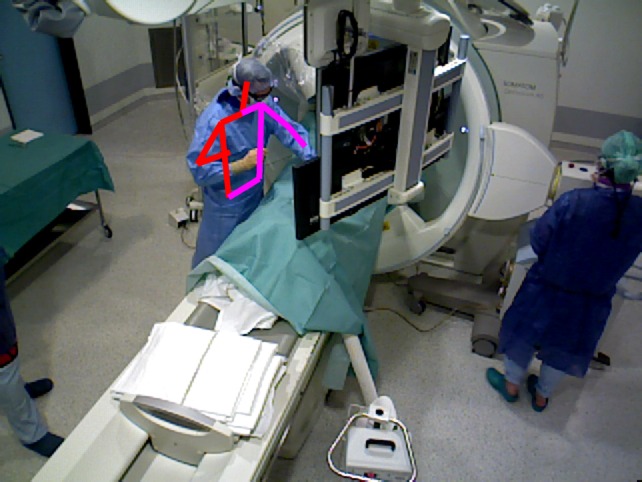} &
\includegraphics[width=0.24\linewidth]{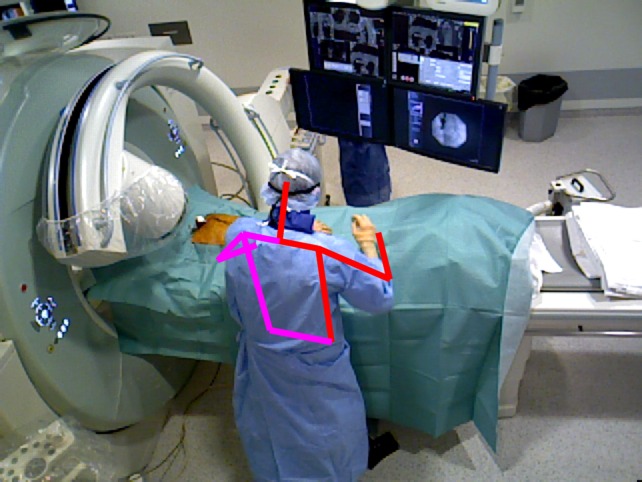} & &
\includegraphics[width=0.24\linewidth]{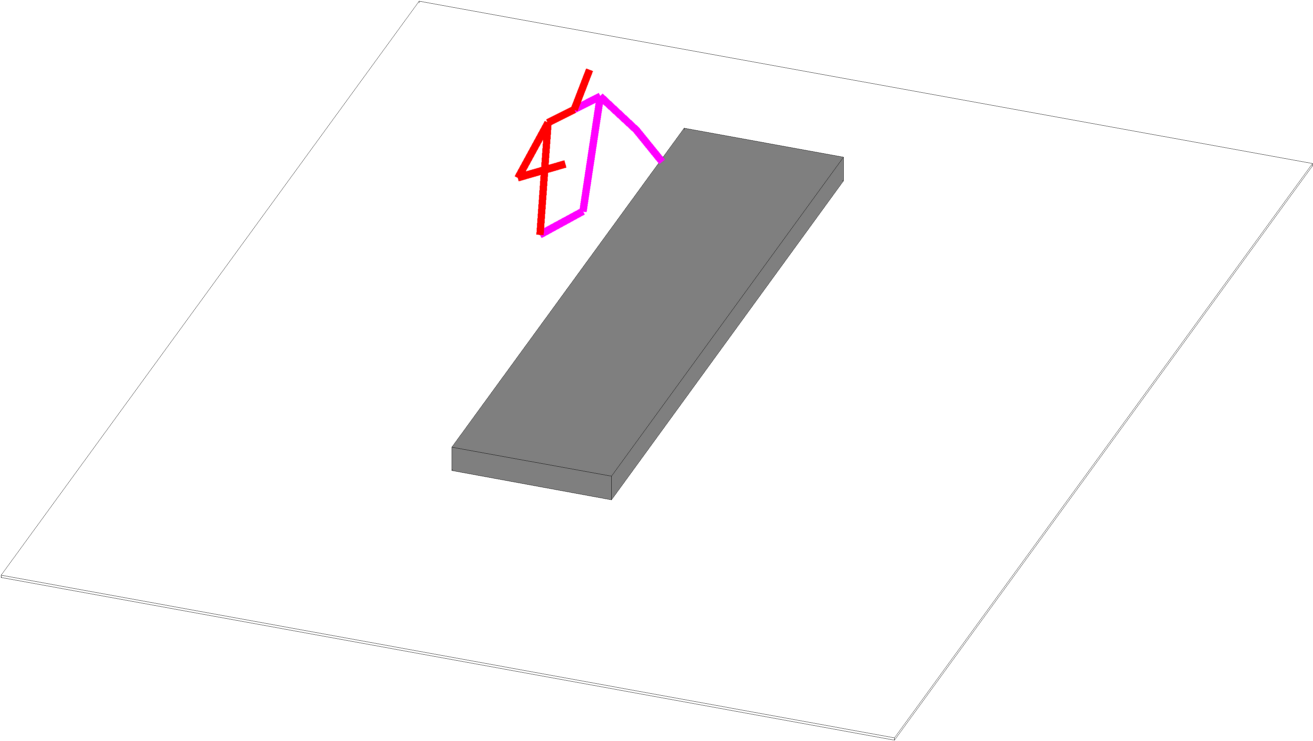} \\ 
\includegraphics[width=0.24\linewidth]{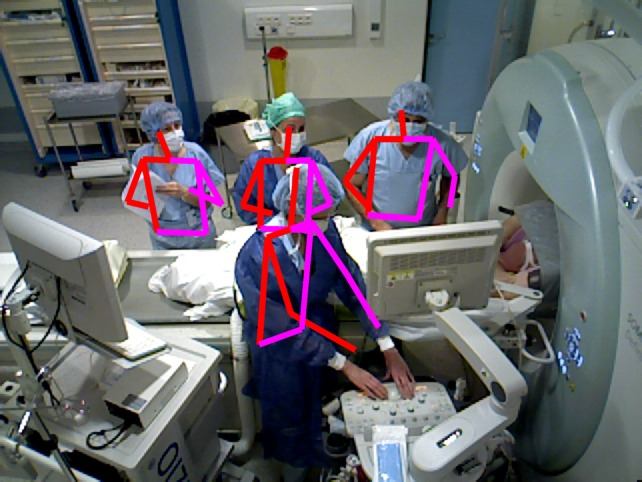} &
\includegraphics[width=0.24\linewidth]{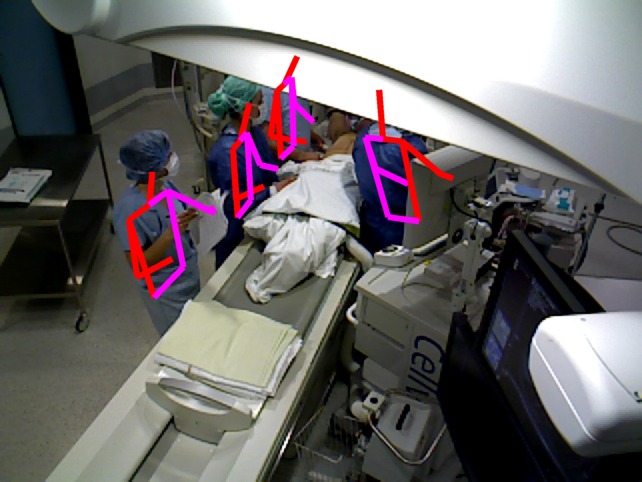} &
\includegraphics[width=0.24\linewidth]{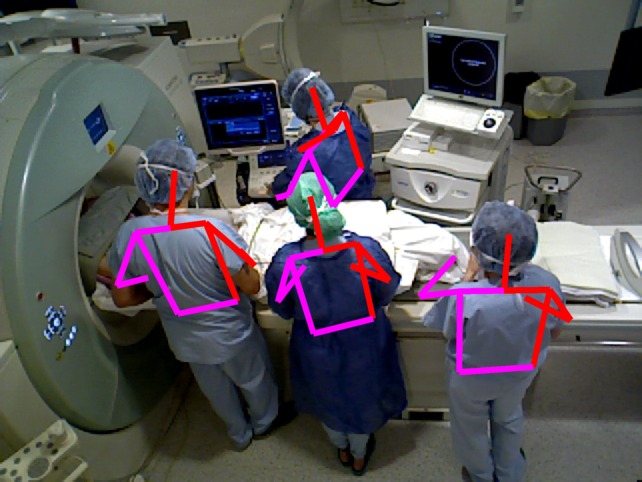} & &
\includegraphics[width=0.24\linewidth]{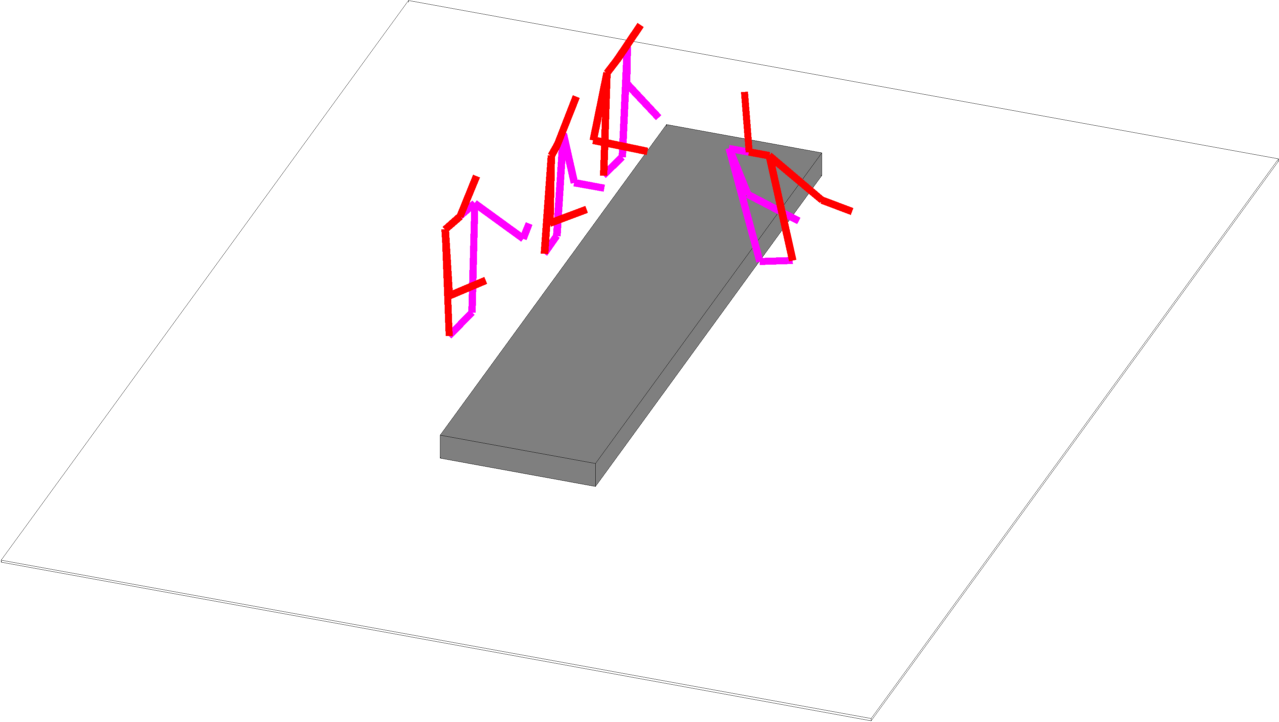} \\  
\includegraphics[width=0.24\linewidth]{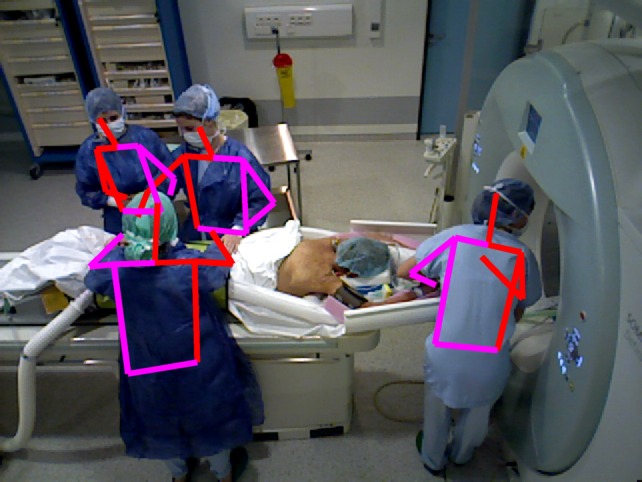} &
\includegraphics[width=0.24\linewidth]{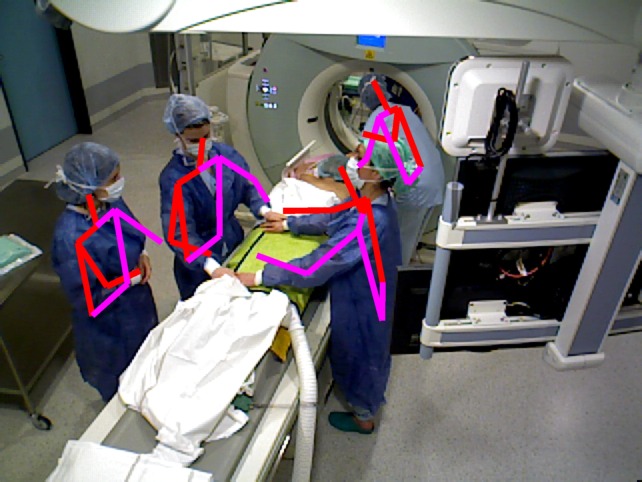} &
\includegraphics[width=0.24\linewidth]{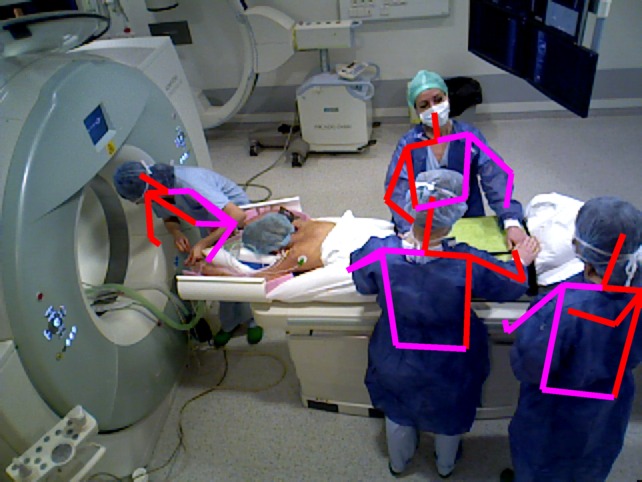} & &
\includegraphics[width=0.24\linewidth]{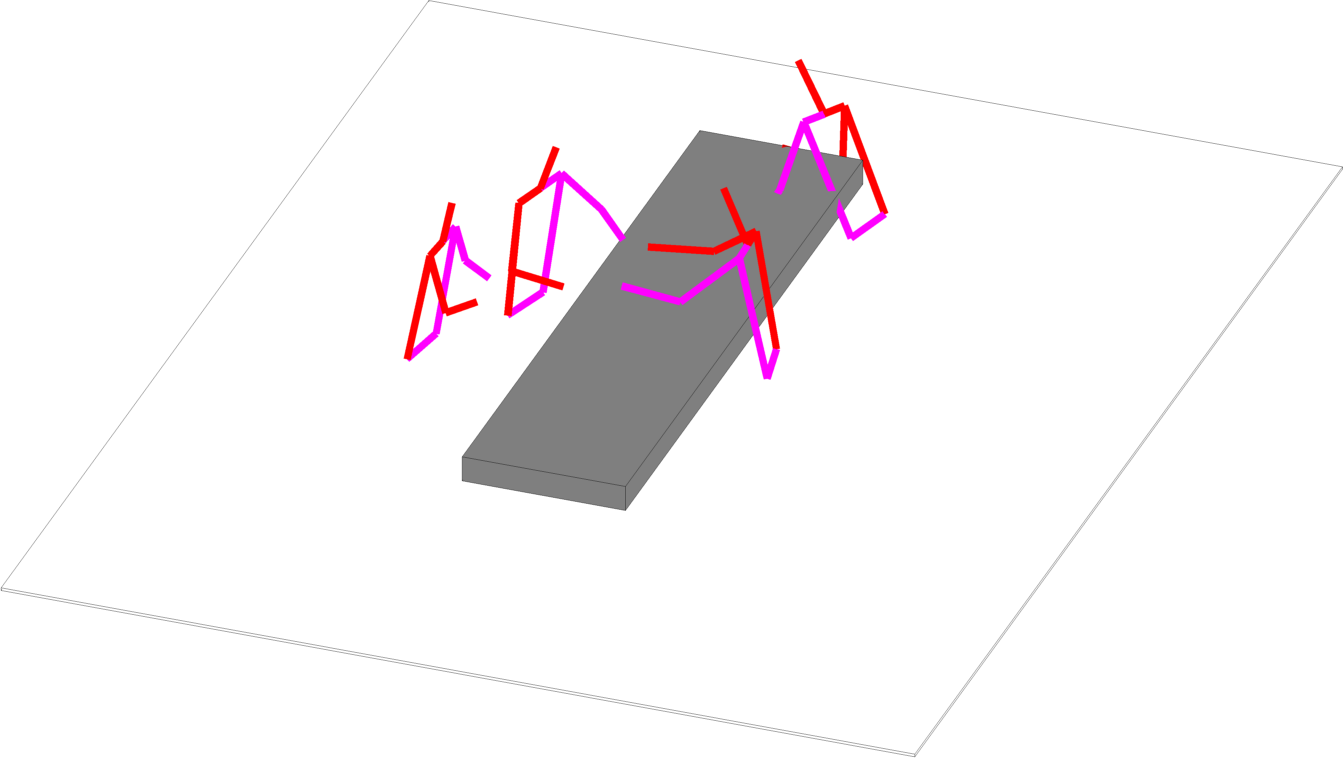}  \\
\includegraphics[width=0.24\linewidth]{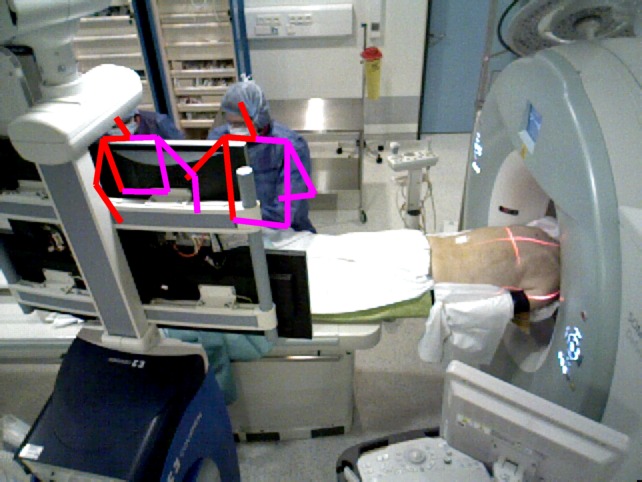} &
\includegraphics[width=0.24\linewidth]{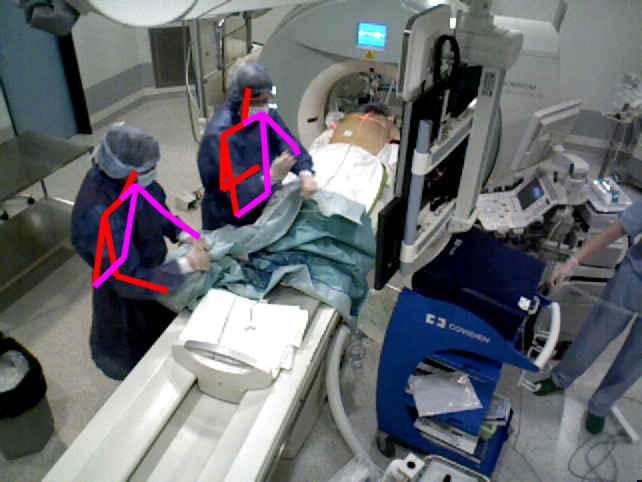} &
\includegraphics[width=0.24\linewidth]{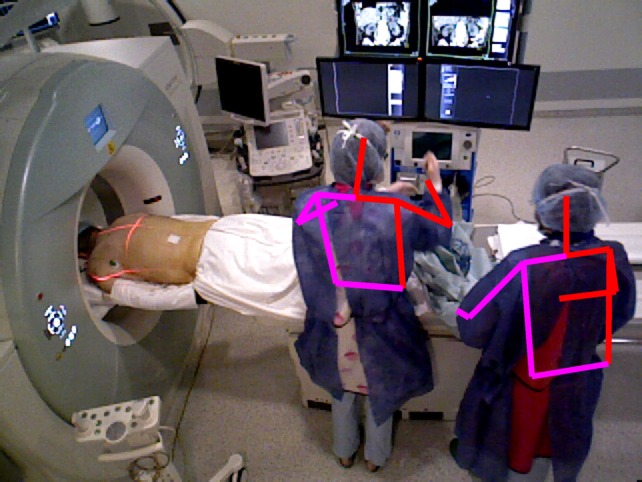} & &
\includegraphics[width=0.24\linewidth]{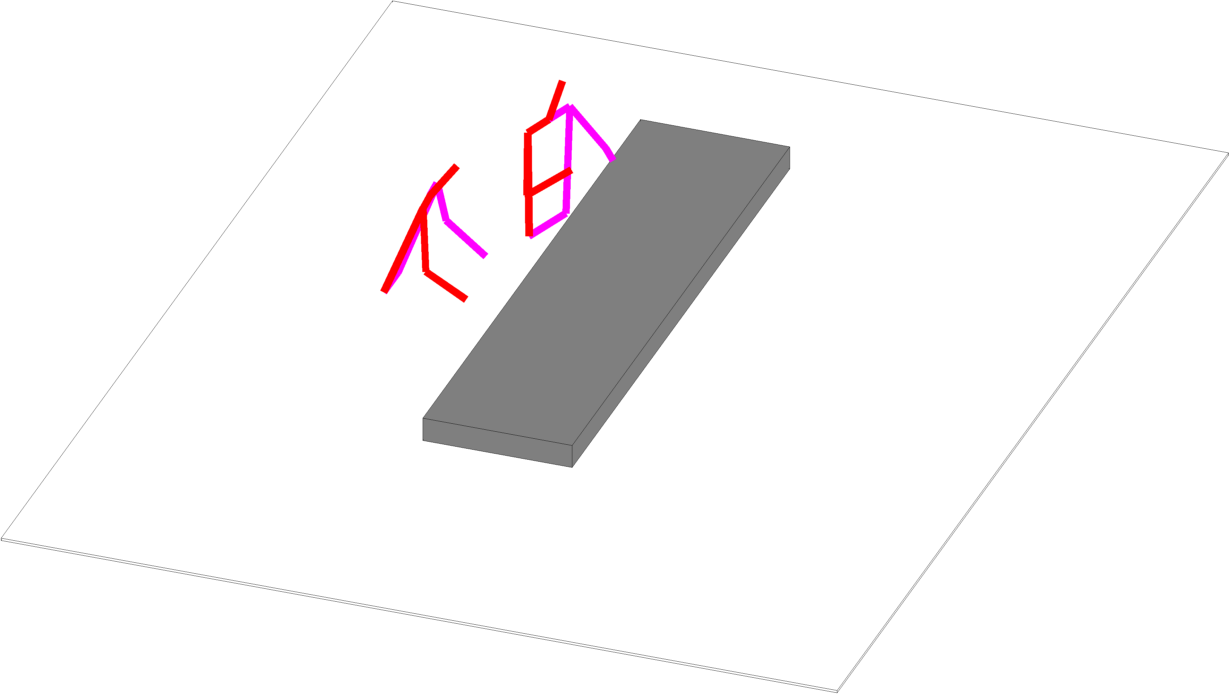} \\
\includegraphics[width=0.24\linewidth]{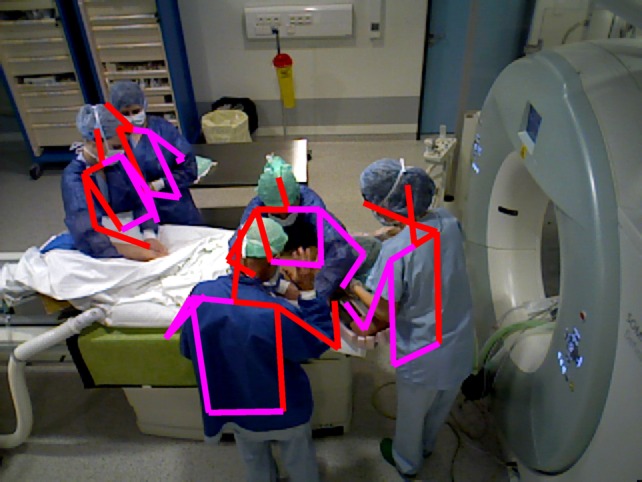} &
\includegraphics[width=0.24\linewidth]{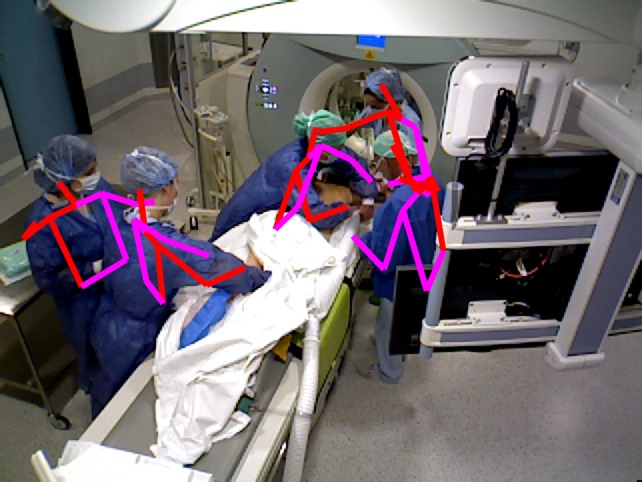} &
\includegraphics[width=0.24\linewidth]{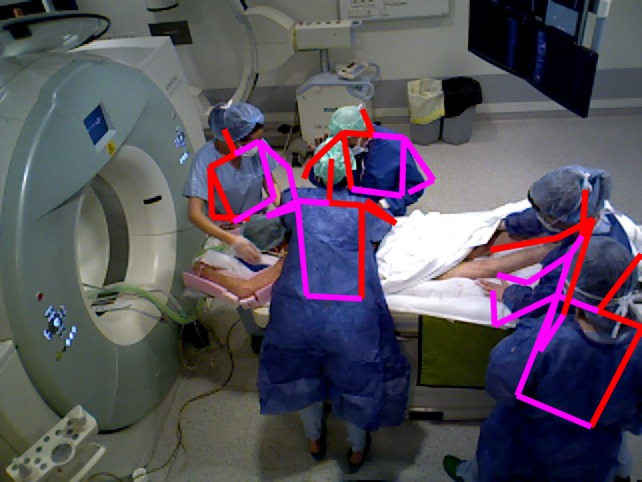} & &
\includegraphics[width=0.24\linewidth]{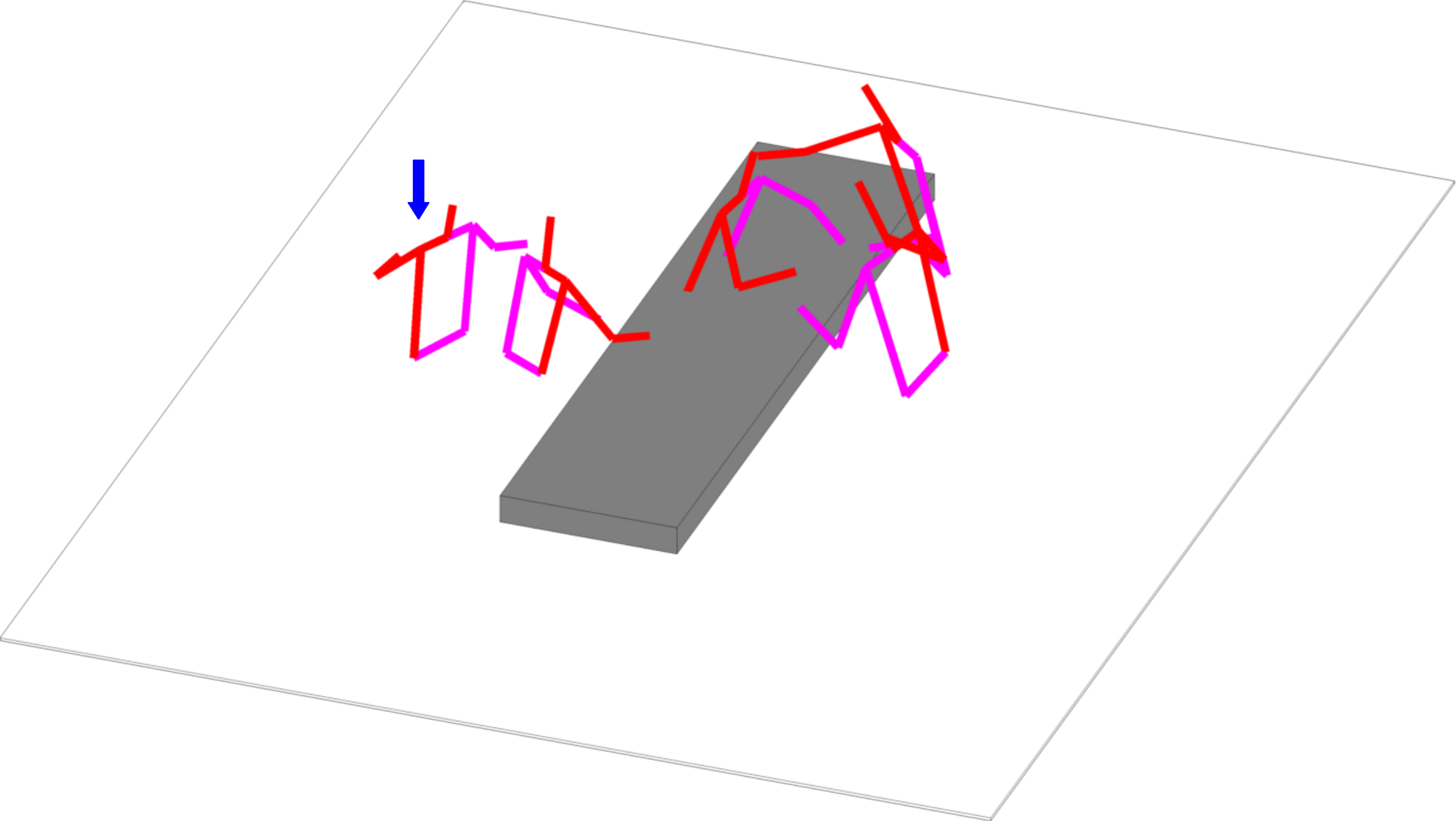} \\
\end{tabular}
\caption{Qualitative results on the multi-view OR dataset. The first three columns in each row show a multi-view frame and the last column shows the corresponding 3D poses for that frame. The overlaid 2D skeletons are computed by projecting 3D poses to the views.  The body parts on the right side of the body are drawn in red. The blue arrow in the last row indicates a physically implausible body pose. (Best viewed in color) }
\label{fig:mvorQual}
\end{figure*}

\subsection{Qualitative Results}

In Figures \ref{fig:h36mQual} and \ref{fig:mvorQual}, we show qualitative results on both Human3.6M and MVOR\footnote{Please note that for generating the qualitative images, the predicted 3D poses are transferred to the room reference frame using an offset computed as the relative difference between the neck location in the ground truth and the neck location in the predicted skeleton.}. Each row shows a multi-view frame. The predicted 3D poses are shown in the last column and the overlaid 2D poses are obtained by projecting the 3D poses into the views.  Figure \ref{fig:h36mQual} demonstrates the high-quality of the predicted 3D body poses.  For example, the frame presented in the last row shows that our approach can successfully incorporate evidence across all views to localize the occluded body parts.  

We also show some frames from the multi-view OR dataset in Figure \ref{fig:mvorQual}. As can be seen in this figure, this dataset is much more complex due to the similar appearance of the objects as well as the people and the presence of many objects and multiple persons in the scenes.  Our approach predicts fairly accurate 3D body poses and always correctly detects the left and right side labels even though it has not seen any data from this dataset or any other data collected in such an OR environment at the training stage\footnote{More qualitative results generated by our model on both datasets are available at \url{https://youtu.be/Cx_kTRzqqzA}}. 

The complexity of this dataset also allows us to identify some of the limitations of the proposed approach. For example, we observe that the elbow and the wrist localization are less accurate compared to other body parts, which is in line with results presented in Tables \ref{tab:2dRes} and \ref{tab:2dResMVOR}. We envision that enforcing appearance consistencies among the projections of a body part across all views can be used to update and improve the 2D body joint detections. The improved 2D detections could then be fed into our multi-view regression model to obtain a more accurate localizations of the body parts in 3D. In the last row of Figure~\ref{fig:mvorQual}, we have highlighted a 3D body pose, where the right arm configuration is infeasible because of body physical constraints. 
We believe that since our training data generation model described in Section \ref{sec:catDet} perturbs 3D poses randomly and does not take the body constraints into account, it may have generated such a training sample. Therefore, it would be interesting to combine our data generation model with a model like the one used in \cite{vondrak_cvpr2008} to enforce and verify the physical plausibility of the generated 3D poses.

\subsection{Ablation study}

\begin{table}[th]
\centering
\setlength{\tabcolsep}{8pt} 
\renewcommand{\arraystretch}{0.9}
\begin{tabular}{lccc}
\toprule
 & EDM & SimpBase & Ours \\
\midrule
GT &  62.2   &  37.1   &  47.2     \\
GT+$\mathcal{N}(0, 5)$ &  67.1 &  46.7 & 48.4  \\
GT+$\mathcal{N}(0,10)$ &  79.1 &  52.8 & 50.8  \\
GT+$\mathcal{N}(0,15)$ &  96.1 &  60.0 & 56.4  \\
GT+$\mathcal{N}(0,20)$ & 115.6 &  70.2 & 65.7\\
\bottomrule
\end{tabular}
\caption{Evaluation results on Human3.6M under noise. Ground truth data is used to both train and test the performance of a variant of our model that relies on single-view input. Similar to EDM \cite{moreno_cvpr2017} and SimpBase \cite{martinez_iccv2017}, we add noise to the test data and gradually increase the amount of noise, where $\mathcal{N}(0,\sigma)$ indicates a normal distribution with mean zero and standard deviation $\sigma$ in pixel.}
\label{tab:noise}
\end{table}

\begin{figure}[!htb]
\centering
\setlength{\tabcolsep}{1pt}
\renewcommand{\arraystretch}{1}
\includegraphics[width=0.96\columnwidth]{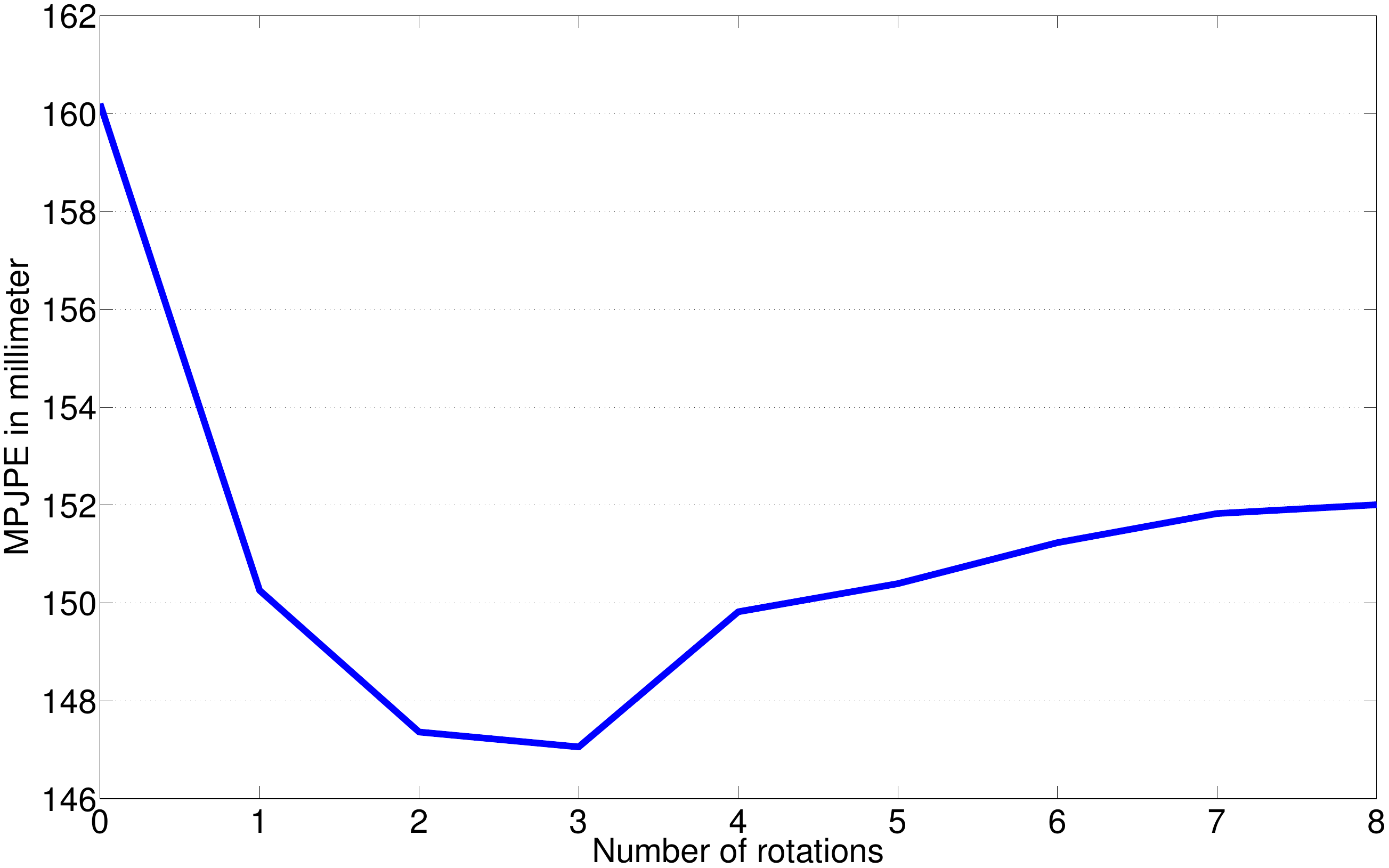} 
\caption{Data augmentation. The results of our multi-view model on MVOR are reported as a function of the number of random rotations of 3D body poses.}
\label{fig:dataAug}
\end{figure}

We performed several experiments on Human3.6M to study the impact of each of the components of our approach. We first observe that by removing the stage-wise supervision, the performance always drops. For example, average MPJPE changes from 57.9 to 77.2 for our $<$MV, Noisy GT$>$ model. Removing batch normalization leads to a substantial increase in the error (from 57.9 to 175). We also observe that the use of dropout during the training of single-view models and multi-view models on perfect ground truth data is important to obtain more robust models, as it reduces the errors by $20 - 50$ mm. However, deactivating dropout for our multi-view models trained on \cite{cao_cvpr2017}'s detections or Noisy GT decreases localization errors by $2$ and $9$ mm, respectively. We believe that this is due to the fact that 2D detection inputs are constructed from single-view poses that have been independently affected by noise in each view by either the detector inaccuracy or by our data generation model. This independent noise can therefore work as a regularizer to enforce neurons to detect the most relevant information across all views, thereby removing the need for dropout.  

Following \cite{moreno_cvpr2017} and \cite{martinez_iccv2017}, we perform a series of experiments to evaluate the performance of our approach under different levels of noise at test time. For a fair comparison, we evaluate our single-view model trained on Noisy GT and add different levels of Gaussian noise to ground truth 2D poses at test time. The evaluation results are presented in Table \ref{tab:noise} and are compared with EDM \cite{moreno_cvpr2017} and SimpBase \cite{martinez_iccv2017}. Even though the average localization error of SimpBase is lower than our model's error by one centimeter when tested on perfect ground truth 2D poses, our model achieves lower localization errors as the noise increases. This indicates that incorporating the detector's characteristics during training allows our model to better cope with the noise at test time.


In a multi-view setup, a 3D body pose can have completely different projections to the views depending on the orientation of the person with respect to the reference coordinate system. We therefore need to construct our multi-view regression model in a way that is robust to these changes in the orientation of the person, as our model only relies on these 2D projections to compute 3D body poses. For this reason, we propose to augment the training data by rotating each 3D pose in human3.6M w.r.t. the reference frame. Figure \ref{fig:dataAug} shows the effect of this data augmentation. We report the results of our multi-view model $<$MV, Noisy GT$>$ on the MVOR dataset as a function of the number of rotations applied to each 3D poses in Human3.6M. 
The results show that applying up to three random rotations decrease the error but applying more random rotation does not lead to any improvement. Apart from the evaluation results reported in Figure \ref{fig:dataAug}, for all the other evaluation on MVOR we always use our multi-view model trained on the train set of Human3.6M, which is augmented by applying three random rotations to each 3D pose.


\section{Conclusions}

We present an easily generalizable approach for estimating 3D body poses using multi-view data. We propose a two-step framework to tackle this problem, which separates single-view pose detection from multi-view 3D pose regression. The proposed approach permits to effectively exploit existing datasets to generalize to new multi-view environments. We use a multi-stage neural network as regression function to estimate 3D poses. Our model is trained on data generated from a set of valid 3D poses by projecting the 3D poses using the camera parameters used at the test time and by incorporating the characteristics of the single-view pose detector. Our evaluation results indicate the effectiveness and importance of incorporating the detector's characteristics during training, as it significantly reduces the localization error and achieves results on par with models trained on the output of the detector. We have also evaluated the generalization of our approach on the multi-person MVOR dataset by using only the camera configuration parameters from this dataset during training, but no image data. Our approach yields fairly accurate results and outperforms the state-of-the-art model on this dataset. The results also show that the localization error dramatically decreases as the number of supporting views increases. This highlights the benefit of our approach in leveraging multi-view data to obtain a reliable model for crowded and cluttered environments. To the best of our knowledge, this is also the first multi-view RGB approach that has been quantitatively evaluated on a real world dataset for the task of 3D body part localization.        

\begin{acknowledgements}

This work was supported by French state funds managed within the Investissements d'Avenir program by BPI France (project CONDOR) and by the ANR (references ANR-11-LABX-
0004, ANR-10-IAHU-02 and ANR-16-CE33-0009). The authors would also like to acknowledge the support of NVIDIA with the donation of a GPU used in this research.

\end{acknowledgements}

\bibliographystyle{plainnat}
\bibliography{refs}

\end{document}